\pdfoutput=1

\documentclass[11pt]{article}

\usepackage{ACL2023}
\usepackage{graphicx}      
\usepackage{amsthm,amsmath,amssymb}
\usepackage[ruled,vlined,boxed]{algorithm2e}
\usepackage{diagbox} 
\usepackage{lipsum} 
\usepackage{algorithmic} 
\usepackage{tcolorbox}
\tcbuselibrary{skins}
\usepackage{longtable}
\usepackage{caption}
\usepackage{multirow}
\usepackage{array}
\usepackage{float} 
\usepackage{booktabs}
\usepackage{enumerate}
\usepackage{enumitem} 
\usepackage{tabularx}
\usepackage{mathrsfs}
\usepackage{makecell}
\usepackage{subcaption}  

\usepackage{times}
\usepackage[T1]{fontenc}

\usepackage[utf8]{inputenc}

\usepackage{microtype}

\usepackage{inconsolata}

\title{M-BRe: Discovering Training Samples for Relation Extraction from Unlabeled Texts with Large Language Models}

\author{Zexuan Li, Hongliang Dai$^{\ast}$, Piji Li \\
\textsuperscript{\rm 1} College of  Artificial Intelligence, \\
Nanjing University of Aeronautics and Astronautics, Nanjing, China\\
\textsuperscript{\rm 2} MIIT Key Laboratory of Pattern Analysis and Machine Intelligence, Nanjing, China\\
\textsuperscript{\rm 3} The Key Laboratory of Brain-Machine Intelligence Technology, Ministry of Education, Nanjing, China.\\
  \texttt{\{zexuanli, hongldai, pjli\}@nuaa.edu.cn} \\}

\begin{document}
\maketitle
\begin{abstract}
For Relation Extraction (RE), the manual annotation of training data may be prohibitively expensive, since the sentences that contain the target relations in texts can be very scarce and difficult to find. It is therefore beneficial to develop an efficient method that can automatically extract training instances from unlabeled texts for training RE models. Recently, large language models (LLMs) have been adopted in various natural language processing tasks, with RE also benefiting from their advances. However, when leveraging LLMs for RE with predefined relation categories, two key challenges arise. First, in a multi-class classification setting, LLMs often struggle to comprehensively capture the semantics of every relation, leading to suboptimal results. Second, although employing binary classification for each relation individually can mitigate this issue, it introduces significant computational overhead, resulting in impractical time complexity for real-world applications. Therefore, this paper proposes a framework called M-BRe to extract training instances from unlabeled texts for RE. It utilizes three modules to combine the advantages of both of the above classification approaches: Relation Grouping, Relation Extraction, and Label Decision. Extensive experiments confirm its superior capability in discovering high-quality training samples from unlabeled texts for RE.
\end{abstract}

\section{Introduction}
Relation Extraction (RE) aims to identify specific relation categories between pairs of entities in texts. It is an essential part of information extraction and has been widely used in knowledge mining \cite{DBLP:journals/csur/ZhongWLPW24,wei2016assessing}, Q\&A systems \citep{DBLP:conf/trec/SrihariL99}, etc. Although existing RE models \citep{DBLP:conf/iclr/PaoliniAKMAASXS21,DBLP:conf/ijcnlp/ZhouC22,DBLP:conf/www/ChenZXDYTHSC22} have performed well on many benchmarks, the scarcity of high-quality training data remains a major problem due to the variety of relation categories in different application domains. Since the sentences that contain the target relation types can be scarce in unlabeled texts, the cost of manually annotating a large training set may become prohibitively expensive when many relation categories are concerned. 


\begin{figure}[t]
\centering
\centerline{\includegraphics[scale=0.35]{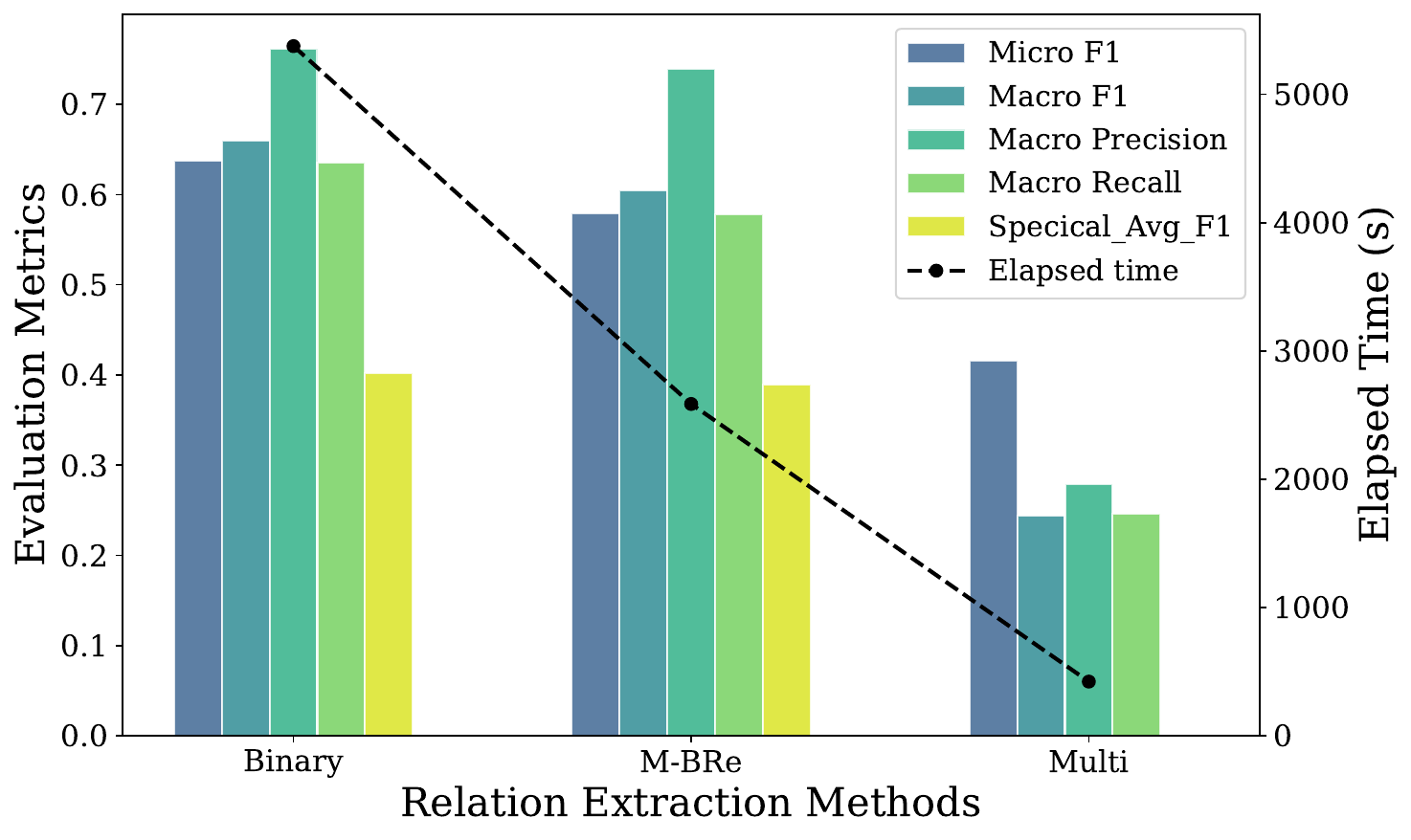}}
\caption{Evaluation metrics across three different Relation Extraction frameworks. The mathematical formulation of each metric is detailed in ``§\ref{Sec:Evaluation} Evaluation''.
}
\label{fig:performance_metrics}
\end{figure}

To address this problem, some studies focus on zero- or few-shot scenarios, and exploit techniques such as meta-learning \citep{DBLP:conf/icml/QuGXT20} and prototypical networks \citep{DBLP:conf/naacl/LiuHWC22}. 
Apart from these techniques,
Large language models (LLMs) have also been employed for implementing RE with limited training data. Two main types of approaches have been investigated. One is to conduct RE directly with LLMs through prompt engineering \citep{DBLP:conf/sustainlp/XuZWZ23}. Another is to use LLMs to generate training examples, which can then be used to learn RE models \citep{DBLP:conf/dasfaa/LiuZGYHLC24, DBLP:conf/acl/LiDL25}. In this paper, we also focus on applying LLMs to automatically produce training data for RE. However, instead of direct generation, we prompt LLMs to discover relation instances from unlabeled texts. This can typically be achieved by calling LLMs to perform a relation classification for each sentence in the texts. However, we observe that directly asking LLMs to conduct multi-class classification would make it difficult for them to understand the semantics of all the relation categories, therefore leading to low-quality prediction results. Another approach is to prompt the model with only one relation type at a time, asking it to perform a binary classification and determine whether this relation exists between the two queried entities in the sentence. However, this approach requires to run the LLM $N$ times for every sentence when there are $N$ target relation categories, which substantially increases the time cost. Figure \ref{fig:performance_metrics} demonstrates the prediction quality and the time cost of these two approaches.


We therefore focus on how to reduce the time cost for LLMs to annotate the unlabeled sentences, while maintaining the correctness.
To this end, we propose the M-BRe framework, which partitions all predefined relation types into multiple groups and ensures that the relations within each group are as distinguishable as possible. For an input example, it performs multi-class classification to differentiate the relations within each group and then further validates each predicted label using binary classification. This two-stage approach enables M-BRe to handle easily distinguishable relations via multi-class classification and more challenging cases via binary classification, thereby combining the strengths of both classification strategies.
As shown in Figure \ref{fig:performance_metrics}, M-BRe achieves comparable performance to binary classification while requiring less than half the running time. 


To comprehensively evaluate our framework, we conducted extensive experiments on standard relation extraction benchmarks, including SemEval and three variants of TACRED. Our results demonstrate that combining the original few-shot manually labeled samples with the framework-generated training samples significantly improves the performance of conventional RE models.

Our main contributions are:

\begin{itemize}
    \item We investigate a novel approach of using LLMs to discover RE training instances from unlabeled texts.

    \item We propose the M-BRe framework, which offers a novel strategy for LLMs to automatically annotate RE labels with both efficiency and accuracy.

    \item We conduct comprehensive experiments to demonstrate the effectiveness of the M-BRe framework and the benefit of the extracted training samples for RE. 
\end{itemize}

Our code is available at \href{https://github.com/Lzx-ZBC/M-BRe}{https://github.com/Lzx-ZBC/M-BRe}.

\begin{figure*}[t]
\centering
\centerline{\includegraphics[scale=0.48]{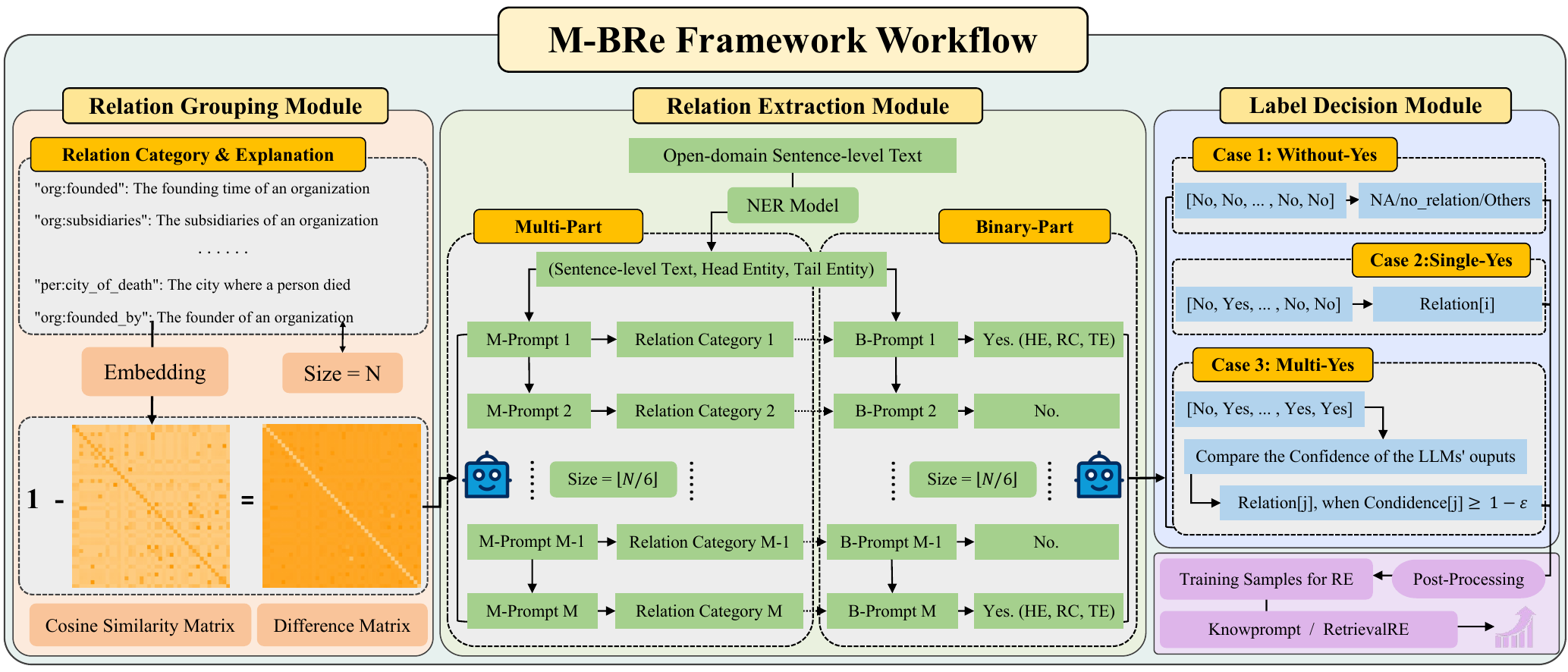}}
\caption{Workflow of the M-BRe Framework, which consists of three modules: Relation Grouping Module, Relation Extraction Module and Label Decision Module.}
\label{fig:B-MRe3}
\end{figure*}

\begin{figure*}[t]
\centering
\centerline{\includegraphics[scale=0.48]{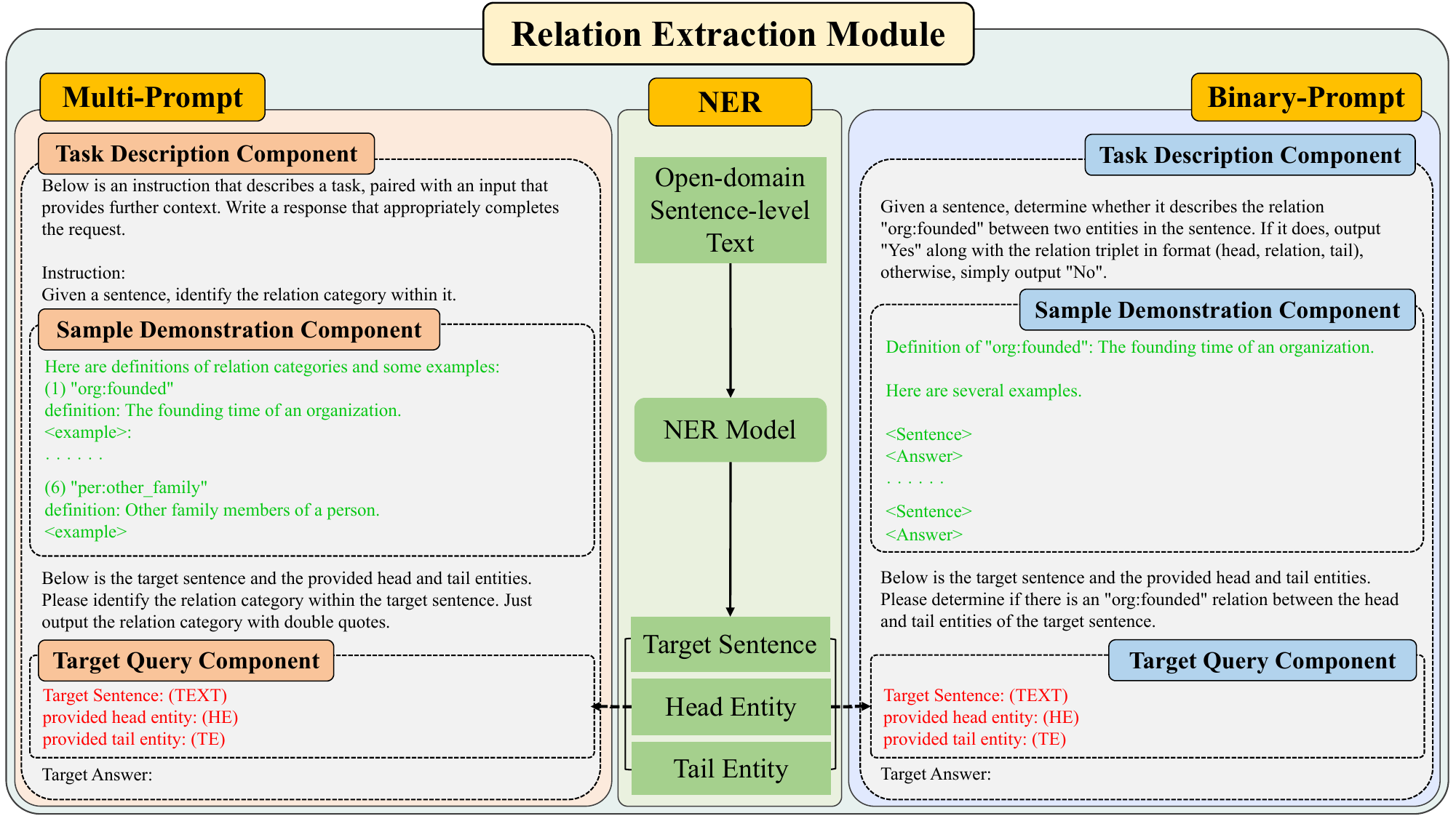}}
\caption{Construction of Multi-Prompt and Binary-Prompt, which consists of three component: Task Description Component, Sample Demonstration Component and Target Query Component.}
\label{fig:prompt}
\end{figure*}

\section{Related Work}
\subsection{Relation Extraction}
Relation Extraction (RE) aims to identify relation categories between head and tail entity pairs in text. As a fundamental natural language processing task, it has been traditionally addressed through machine learning approaches such as Bootstrap and Snowball \citep{DBLP:conf/emnlp/BatistaMS15,DBLP:conf/aaai/GaoHX0LLS20}. Subsequent advancements in deep learning led to the adoption of pipeline architectures employing CNN, RNN and LSTM \citep{DBLP:conf/emnlp/ZengLC015,DBLP:conf/acl/MiwaB16,DBLP:conf/ijcnlp/ZhouC22,DBLP:conf/emnlp/ZhangZCAM17}, along with joint End2End framework and graph neural network model \citep{DBLP:conf/emnlp/Zhang0M18,DBLP:conf/acl/GuoZL19,DBLP:conf/ijcai/GuoN0C20} for RE. Since the emergence of Pre-trained Language Models (PLMs) like BERT, PLMs-based RE models \citep{DBLP:conf/naacl/DevlinCLT19,DBLP:conf/nlpcc/HuangCWC19,DBLP:conf/ijcnn/MoreiraOMZB20,DBLP:conf/www/ChenZXDYTHSC22} have become the dominant paradigm due to their superior performance. 

Recent research has demonstrated growing interest in employing Large Language Models (LLMs) for direct RE. \citet{DBLP:conf/sustainlp/XuZWZ23} introduced a prompting strategy that incorporates comprehensive relation category definitions and annotated examples, enabling LLMs to better comprehend RE task specifications. Their empirical results confirm that LLMs can generate highly accurate RE predictions. \citet{DBLP:conf/acl/ZhangG023} developed QA4RE, a novel framework that formalizes RE as a Question Answering (QA) task. Additionally, \citet{DBLP:conf/emnlp/LiWK23} proposed SUMASK, an advanced prompting technique that reformulates RE through task decomposition into text summarization and QA components, thereby enhancing LLMs' compatibility.
\subsection{Data Generation}
The scarcity of high-quality data has long been a key factor constraining model performance, making data generation an emerging hotspot. There were already some studies on this topic before instruction tuned LLMs become popular \cite{DBLP:conf/emnlp/YeGLXF00K22,DBLP:conf/nips/MengHZH22,DBLP:conf/iclr/GaoPLXY0ZLLK23}. \citet{DBLP:conf/nips/MengHZH22} prompts the PLMs to create training data for natural language understanding tasks. \citet{DBLP:conf/acl/ChiaBPS22} presents a framework that leverages language models to generate structured text for synthesizing unseen relation categories. Recently LLMs have demonstrated strong language generation capabilities, providing a viable alternative for synthetic data generation \citep{DBLP:conf/acl/HartvigsenGPSRK22,DBLP:conf/acl-convai/SahuRLAVB22}. \citet{DBLP:conf/sustainlp/XuZWZ23} use the LLMs to generate data for assisting the models themselves on RE. \citet{DBLP:conf/acl/LongWXZDCW24} pointed out that effective prompts consist of three key elements: Task Specification, Generation Conditions, and In-Context Demonstrations. By clarifying these elements, the accuracy and diversity of generated data can be significantly improved. For complex data generation tasks, Multi-Step Generation has emerged as an important strategy. By breaking down the generation process into multiple simpler subtasks, complex data structures can be generated step by step. 

These results show that using LLMs to generate data can effectively enhance the performance of RE models. However, they primarily focus on leveraging the intrinsic knowledge base of LLMs, without attempting to construct data from large-scale real-world unlabeled texts.

\section{The M-BRe Framework}
In this section, we present the architectural details of the M-BRe framework, which comprises three key modules as illustrated in Figure \ref{fig:B-MRe3}: Relation Grouping, Relation Extraction, and Label Decision. 

The framework starts by partitioning all the predefined relation types into $K$ groups with \textit{Relation Grouping Module}, where the relations in each group should be as distinguishable as possible.
Consequently, it decomposes the single large-scale multi-class classification task into $K$ smaller-scale multi-class classification subtasks.
For an input example, each subtask is completed with the multi-class classification prompt of \textit{Relation Extraction Module}, thereby yielding $K$ relation labels. To further select from these $K$ labels, we consider each of them individually, and leverage the binary classification prompt of \textit{Relation Extraction Module} $K$ times to infer whether each of them is a valid label. Finally, \textit{Label Decision Module} is used to determine the final labels for the input example.

\subsection{Relation Grouping Module}
\label{section:Relation Grouping Module}

We first construct an explanation for each predefined relation category in the RE dataset that follows the form ``"org:founded": The founding time of an organization.'' The details of the explanations for all relation categories are provided in Appendix \ref{sub:RES}. We then vectorize them and compute their cosine similarities, deriving a matrix that quantifies the semantic disparities between different relation labels. To partition the relations based on this matrix, the two most dissimilar relation categories are used as initial seed groups. Then, the remaining relations are iteratively assigned to the group that minimizes the maximum similarity of the group using a greedy strategy, while keeping the size of each group balanced. The pseudo code for this process is shown in Appendix \ref{algorithm:RG}. 


Given N predefined relation categories, we set the number of groups to $\lfloor N/6 \rfloor$. This grouping strategy and its empirical validation are systematically analyzed in ``§\ref{section:Number of Groupings} Number of Groups''.

\subsection{Relation Extraction Module}
\label{section:Relation Extraction Module}
This module focuses on the prompt design for both multi-class classification and binary classification. Both prompts consist of three fundamental components: Task Description, Sample Demonstration, and Target Query. 
The details of the prompts are shown in Figure \ref{fig:prompt}.

\paragraph{Task Description Component.} This component consists of two basic parts. At the beginning of the prompt, we instruct the LLM that the objective is to identify the entity relation within a sentence. Then, following a few demonstrations, the model is told that the target sentence along with given head-tail entity pairs will be provided. The output format is also explicitly specified. In multi-class classification settings, the LLMs are required to predict the relation category directly. For binary classification, the LLMs are expected to output ``Yes. (Head Entity, Relation Category, Tail Entity)'' if they think that the current relation category does exist in the input unlabeled sentence, otherwise they should output ``No.''.


\paragraph{Sample Demonstration Component.}
We use each grouping result obtained in ``§\ref{section:Relation Grouping Module} Relation Grouping Module'' to construct the multi-class classification prompt, and employ In-Context Learning (ICL) \citep{DBLP:conf/nips/BrownMRSKDNSSAA20,DBLP:conf/emnlp/WanCMLSLK23} to provide definition and demonstrations for each relation category in the group, so that LLMs can distinguish the relation types within the group more easily. 
The binary classification prompt also applies ICL to provide 3 correct and 4 incorrect examples for the current relation category. 

\paragraph{Target Query Component.}
We extract all entities from unlabeled sentence-level text and randomly sample two of them to form a head-tail entity pair. Finally, it can be injected into the prompt, which will enhance the LLMs' ability to discern potential relation categories.

\subsection{Label Decision Module}
\label{section:Confidence Judgment Module}
With ``§\ref{section:Relation Extraction Module} Relation Extraction Module'', we will get $\lfloor N/6 \rfloor$ ``Yes'' or ``No'' results. In order to determine the final relation extraction results, a Confidence-based label decision strategy is introduced to handle the following three scenarios that may occur. The confidence for LLM-generated content is formulated as follows:
\begin{equation*}
    \mathrm{Confidence}= \mathrm{\frac{1}{M}\sum_{i=1}^M\max(softmax(logits_i))},
\end{equation*}
where $\mathrm{M}$ is the number of tokens generated.

\paragraph{Case 1: Without-Yes.}
The result of binary classification are all ``No.'', indicating that the LLMs consider that the unlabeled text does not have any relation categories, so we set the relation category between the head-tail entity pairs of the current unlabeled text to ``NA'' or ``no\_relation'' or ``Others”.
\paragraph{Case 2: Single-Yes.}
The result of the binary classification is only one ``Yes. (Head Entity, Relation Category, Tail Entity)'', indicating that the LLMs consider that one of $\lfloor N/6 \rfloor$ relation categories, $R_1$, exists in the current unlabeled text, and we set the relation category between the head-tail entity pairs of the current unlabeled text to $R_1$.
\paragraph{Case 3: Multi-Yes.}
There are several ``Yes. (Head Entity, Relation Category, Tail Entity)'' results of the binary classification, indicating that the LLMs consider that there are more than one of $\lfloor N/6 \rfloor$ relation categories in the current unlabeled text, $\mathrm{[R_1,R_2, ... ,R_i]}$. We calculate the confidence of LLMs' binary classification predictions for each relation category and set the relation categories between the head-tail entity pairs of the current unlabeled text to $\mathrm{[R_1,R_2, ...,R_j]\;(j\leq i)}$, where $\mathrm{Confidence\;(R_j)\geq 1- \epsilon,\;\epsilon=10^{-2}}$. 

In our current implementation, we set the confidence threshold to a fixed value $\theta = 10^{-2}$ based on preliminary experiments and empirical observations, which showed that this value provided a good balance between precision and recall in our label decision process.

\begin{itemize}
    \item If we significantly reduce the $1-\theta$ value dynamically, the subsequently constructed training samples would include relation categories less suitable for the current entity pairs. The more the $1-\theta$ value decreases, the greater the noise in the training samples regarding relation categories.

    \item If we only make minimal dynamic adjustments to the $1-\theta$ value, the results would be close to those obtained with the current $1-\theta$ setting.
\end{itemize}

\begin{table}[t]

\centering
\renewcommand{\arraystretch}{1}
\scalebox{0.73}{
\begin{tabular}{cccc|ccc}
  \toprule
   Dataset & \multicolumn{3}{c}{Tacred} & \multicolumn{3}{c}{SemEval}\\
   \cmidrule(r){1-7}
  \multirow{2}*{$\theta$} & \multicolumn{3}{c}{Qwen2.5-7B-Instruct-1M} & \multicolumn{3}{c}{Qwen2.5-7B-Instruct-1M} \\
  \cmidrule(r){2-4}\cmidrule(r){5-7}
  & Mi-F1 & Ma-F1 & \multicolumn{1}{c|}{S\_A\_F1}
  & Mi-F1 & Ma-F1 & S\_A\_F1 \\
  \cmidrule(r){1-7}
  \multicolumn{1}{c|}{0.001} & 55.07&	58.80	&36.28&	34.07&	31.73	&25.64\\
  \multicolumn{1}{c|}{0.01} & 53.14&	56.08&	35.05&	43.66&	43.36	&33.68\\
  \multicolumn{1}{c|}{0.02} & 56.04&	60.28	&33.65	&47.25	&46.59&	35.53 \\
  \multicolumn{1}{c|}{0.05} & 59.90&	61.98	&32.72	&43.96	&42.72	&28.57\\
  \multicolumn{1}{c|}{0.1} & 58.94	&60.16&	33.17	&36.26&	34.91	&23.44\\
  \cmidrule(r){1-7}
  \multirow{2}*{$\theta$} & \multicolumn{3}{c}{Qwen2.5-14B-Instruct-1M} & \multicolumn{3}{c}{Qwen2.5-14B-Instruct-1M} \\
  \cmidrule(r){2-4}\cmidrule(r){5-7}
  & Mi-F1 & Ma-F1 & \multicolumn{1}{c|}{S\_A\_F1}
  & Mi-F1 & Ma-F1 & S\_A\_F1 \\
  \cmidrule(r){1-7}
  \multicolumn{1}{c|}{0.001} & 59.90	&63.71	&34.37	&52.75	&51.77	&37.55\\
  \multicolumn{1}{c|}{0.01} & 58.94&	63.95&	33.86	&52.75&	51.14	&38.64\\
  \multicolumn{1}{c|}{0.02} & 58.12	&62.48&	33.93&	54.95	&52.95&	36.63 \\
  \multicolumn{1}{c|}{0.05} & 59.08&	62.55	&30.76	&56.04	&56.47	&34.80\\
  \multicolumn{1}{c|}{0.1} & 61.50&	65.08&	31.39	&54.95	&54.79	&36.26\\
  \cmidrule(r){1-7}
  \multirow{2}*{$\theta$} & \multicolumn{3}{c}{Qwen3-14B} & \multicolumn{3}{c}{Qwen3-14B} \\
  \cmidrule(r){2-4}\cmidrule(r){5-7}
  & Mi-F1 & Ma-F1 & \multicolumn{1}{c|}{S\_A\_F1}
  & Mi-F1 & Ma-F1 & S\_A\_F1 \\
  \cmidrule(r){1-7}
  \multicolumn{1}{c|}{0.001} & 75.36&	76.72	&40.75&	20.88&	20.88	&16.67\\
  \multicolumn{1}{c|}{0.01} & 76.33&	77.46	&39.58&	21.98&	22.28	&16.85\\
  \multicolumn{1}{c|}{0.02} & 76.33	&77.81	&37.79	&19.78	&19.12	&15.56 \\
  \multicolumn{1}{c|}{0.05} & 76.33	&77.05&	36.51	&19.78	&19.16	&14.84\\
  \multicolumn{1}{c|}{0.1} & 77.78&	78.97&	37.60	&17.58&	16.05	&12.27\\
  \bottomrule
\end{tabular}}
\caption{\label{tab:theta} Performance results of the M-BRe framework under different threshold $\theta$. Mi-F1, Ma-F1 and S\_A\_F1 represent Micro F1, Macro F1 and Special\_Avg\_F1.
}
\end{table}

To give a better visualization of this phenomenon, we dynamically adjusted $\theta$ to take values in [0.001, 0.01, 0.02, 0.05, 0,1] and the experimental results are shown in Tabel \ref{tab:theta}. It can be seen that, in most settings, variations of $\theta$ in the range of [0.001, 0.01, 0.02, 0.05, 0,1] do not have a sensitive effect on the M-BRe framework, especially when 14B models are used. In general, our setting of $\theta = 10^{-2}$ allows the approach to achieve good performance results in most experimental setups.


\begin{table*}[h]
\centering
\renewcommand{\arraystretch}{1}
\scalebox{0.84}{
\begin{tabular}{cc|ccc|ccc|ccc|ccc}
  \toprule
  \multicolumn{2}{c}{\multirow{2}{*}{\diagbox{Method}{Dataset}}} & \multicolumn{3}{c}{TACRED} & \multicolumn{3}{c}{TACRED-Revisit} & \multicolumn{3}{c}{Re-TACRED} & \multicolumn{3}{c}{SemEval}\\
  \cmidrule(r){3-5}\cmidrule(r){6-8}\cmidrule(r){9-11}\cmidrule(r){12-14}
  \multicolumn{2}{c}{} & K=2 & K=4 & K=8 & K=2 & K=4 & K=8 & K=2 & K=4 & K=8 & K=2 & K=4 & K=8 \\
  \midrule
  \multirow{6}{*}{\rotatebox[origin=c]{90}{\parbox{3cm}{\centering\textbf{Qwen2.5-7B-Instruct-1M}}}} 
  & Knowprompt & 5.91 & 11.24 & 22.07 & 12.44 & 14.04 & 26.31 & 12.35 & 14.48 & 44.77 & 15.71 & 38.18 & 61.39\\  
  & Mix 4 & 14.95 & 18.73 & 28.20 & 19.72 & 20.85 & 27.36 & 20.89 & 34.73 & 39.34 & 32.97 & 43.74 & 66.98 \\
  & Mix P & \textbf{20.32} & \textbf{25.36} & \textbf{30.13} & \textbf{24.52} & \textbf{25.59} & \textbf{28.96} & \textbf{27.48} & \textbf{35.07} & \textbf{53.25} & \textbf{37.81} & \textbf{44.71} & \textbf{68.13} \\	
  \cmidrule(r){2-14}
  & RetrievalRE & 15.34 & 13.34 & 28.14 & 17.35 & 19.53 & 28.95 & 9.76 & 17.54 & 33.72 & 38.95 & 52.92 & 72.53 \\  
  & Mix 4 & 16.71 & 20.56 & 27.25 & 21.18 & 26.56 & 29.61 & 18.10 & 25.19 & 35.13 & 40.12 & 53.27 & 68.94 \\
  & Mix P & \textbf{20.44} & \textbf{25.37} & \textbf{30.30} & \textbf{22.25} & \textbf{26.84} & \textbf{31.27} & \textbf{21.72} & \textbf{26.44} & \textbf{36.30} & \textbf{42.71} & \textbf{56.89} & \textbf{73.33}\\
  \midrule
  \multirow{6}{*}{\rotatebox[origin=c]{90}{\parbox{3cm}{\centering\textbf{Qwen2.5-14B-Instruct-1M}}}} 
  & Knowprompt & 5.91 & 11.24 & 22.07 & 12.44 & 14.04 & 26.31  & 12.35 & 14.48 & 44.77 & 15.71 & 38.18 & 61.39 \\
  & Mix 4 & 12.26 & 19.30 & 25.65 & 14.55 & 19.18 & 26.06 & 19.03 & 26.55 & 36.28 & 34.28 & 38.33 & 45.18 \\
  & Mix P & \textbf{16.76} & \textbf{21.94} & \textbf{27.05} & \textbf{17.44} & \textbf{21.41} & \textbf{30.00} & \textbf{20.20} & \textbf{41.86} & \textbf{50.97} & \textbf{44.88} & \textbf{54.37} & \textbf{71.38} \\
  \cmidrule(r){2-14}
  & RetrievalRE & 15.34 & 13.34 & 28.14 & 17.35 & 19.53 & 28.95 & 9.76 & 17.54 & 33.72 & 38.95 & 52.92 & 72.53 \\ 
  & Mix 4 & 17.19 & 21.94 & 26.71 & 19.40 & 23.34 & 30.02 & 20.78 & 25.16 & 39.00 & 50.77 & 59.01 & 64.60 \\
  & Mix P & \textbf{18.80} & \textbf{23.78} & \textbf{28.91} & \textbf{21.51} & \textbf{26.48} & \textbf{30.25} & \textbf{21.24} & \textbf{25.84} & \textbf{40.79} & \textbf{56.09} & \textbf{61.00} & \textbf{76.20} \\
  \midrule
  \multirow{6}{*}{\rotatebox[origin=c]{90}{\parbox{3cm}{\centering\textbf{Qwen3-14B}}}} 
  & Knowprompt & 5.91 & 11.24 & 22.07 & 12.44 & 14.04 & 26.31  & 12.35 & 14.48 & 44.77 & 15.71 & 38.18 & 61.39 \\
  & Mix 4 & 19.14 & 25.34 & 29.24 & 23.30 & 27.01 & 30.10 & 22.67 & 33.46 & 48.25 & 31.43 & 33.01 & 53.98\\  
  & Mix P & \textbf{21.14} & \textbf{25.96} & \textbf{30.12} & \textbf{24.77} & \textbf{29.22} & \textbf{31.23} & \textbf{27.66} & \textbf{38.84} & \textbf{51.15} & \textbf{35.44} & \textbf{45.89} & \textbf{62.67}\\  
  \cmidrule(r){2-14}
  & RetrievalRE & 15.34 & 13.34 & 28.14 & 17.35 & 19.53 & 28.95 & 9.76 & 17.54 & 33.72 & 38.95 & 52.92 & 72.53 \\ 
  & Mix 4 & 16.34 & 21.82 & 27.81 & 21.04 & 21.78 & 29.46 & 18.39 & 23.86 & 40.02 & 40.98 & 53.41 & 66.76\\  
  & Mix P & \textbf{19.73} & \textbf{22.14} & \textbf{28.64} & \textbf{21.54} & \textbf{25.69} & \textbf{29.72} & \textbf{21.19} & \textbf{31.52} & \textbf{47.46} & \textbf{41.18} & \textbf{58.54} & \textbf{72.54}\\  
  \bottomrule
\end{tabular}}
\caption{\label{main-result}
Micro F1 (\%) of few-shot performance. \textbf{Knowprompt} and \textbf{RetrievalRE} mean the performance of manually labeled training samples only. \textbf{Mix 4} and \textbf{Mix P} mean the performance of combining the use of manually labeled training samples and constructed training samples when relation categories are divided into 4 and P groups, where P =
$\lfloor N/6 \rfloor$ and $N$ is the total number of relation categories for each RE dataset.
}
\end{table*}

\begin{table}[ht] 

\centering
\renewcommand{\arraystretch}{0.95}
\scalebox{0.86}{
\begin{tabular}{@{}lccccc@{}} 
\toprule
LLMs & Model         & Tacred           & Tacrev           & Retacred          & SemEval      \\
\midrule

\multirow{4}{*}{{\parbox{1.1cm}{\textbf{Qwen2.5-7B-Instruct-1M}}}} &
\multirow{2}{*}{Kp.} & 9.80 & 11.61 & 14.03 & 19.15\\
 & & \textbf{17.52} & \textbf{19.03} & \textbf{20.36} & \textbf{19.70}         \\
 \cmidrule(r){3-6}
 & \multirow{2}{*}{Rt.}& 13.23 & 15.10 & 13.82 & 18.97\\
 & &\textbf{16.57} & \textbf{20.60} & \textbf{17.53} & \textbf{19.46}         \\
\midrule 
\multirow{4}{*}{{\parbox{1.1cm}{\textbf{Qwen2.5-14B-Instruct-1M}}}} &
\multirow{2}{*}{Kp.} & 9.58 & 11.43 & 11.34 & 28.99\\
 & & \textbf{9.64} & \textbf{15.60} & \textbf{15.31} & \textbf{37.64}         \\
 \cmidrule(r){3-6}
&\multirow{2}{*}{Rt.}& 12.92 & 13.86 & 12.04 & 31.98\\
 & & \textbf{17.69} & \textbf{16.22} & \textbf{14.13} & \textbf{32.77}         \\
\midrule 
\multirow{4}{*}{{\parbox{1.1cm}{\textbf{Qwen3-14B}}}} &
\multirow{2}{*}{Kp.} & 13.42 & 14.69 & 13.68 & 11.88\\
 & &\textbf{17.99} & \textbf{16.67} & \textbf{19.54} & \textbf{26.53}         \\
 \cmidrule(r){3-6}
&\multirow{2}{*}{Rt.}& 12.50 & 11.77 & 9.77 & 11.56\\
 & &\textbf{16.53} & \textbf{16.38} & \textbf{15.13} & \textbf{29.99}         \\
\bottomrule
\end{tabular}}

\caption{\label{tab:pure}Micro F1 (\%) of few-shot performance using constructed training samples only. The first and second line of each model correspond to \textbf{Pure 4} and \textbf{Pure P}.}
\end{table}

\section{Experimental Settings}
\subsection{Datasets}
We conducted experiments on SemEval and three versions of TACRED: SemEval 2010 Task 8 (SemEval) \citep{DBLP:conf/semeval/HendrickxKKNSPP10}, TACRED \citep{DBLP:conf/emnlp/ZhangZCAM17}, TACRED-Revisit \citep{DBLP:conf/acl/AltGH20a}, Re-TACRED \citep{DBLP:conf/aaai/StoicaPP21}. Statistical details are given in Appendix \ref{appendix:datasets}.

\subsection{Evaluation}
\label{Sec:Evaluation}
\paragraph{For M-BRe Framework.} We randomly select 5 test samples for each relation category from the TACRED and SemEval test datasets. Since some relation categories have fewer than 5 test samples, the final number of samples used for testing is 207 for TACRED and 91 for SemEval. As long as the predicted relation list includes the ground truth, the prediction of the M-BRe framework is considered correct. In this way, Micro F1, Macro F1, Macro precision, Macro recall, and Elapsed time are used as preliminary evaluation metrics. Meanwhile, in order to reflect the absoluteness of the correct prediction of the M-BRe framework, we introduce the Special\_Avg\_F1 metric, which takes into account the length of the predicted relation list. The formula is specified as follows:

\begin{equation*}
    \mathrm{Precison_i}=\mathrm{\frac{|\cap(PredictionSet_i, ReferenceSet_i)|}{|PredictionSet_i|}},
\end{equation*}
\begin{equation*}
    \mathrm{Recall_i}=\mathrm{\frac{|\cap(PredictionSet_i, ReferenceSet_i)|}{|ReferenceSet_i|}},
\end{equation*}
\begin{equation*}
    \mathrm{Special\_Avg\_F1}
    =\mathrm{\frac{1}{N}}\mathrm{\sum_{i=0}^{N}} \mathrm{\frac{2\times Precision_i\times Recall_i}{Precision_i+ Recall_i}},
\end{equation*}

where for each test sample $\mathrm{i}$, $\mathrm{Precision_i}$ and $\mathrm{Recall_i}$ take $\mathrm{10^{-10}}$ only if the intersection of $\mathrm{PredictionSet_i}$ and $\mathrm{ReferenceSet_i}$ is $\varnothing$.

\begin{table*}[h]

\centering
\renewcommand{\arraystretch}{1}
\scalebox{0.86}{
\begin{tabular}{cc|ccc|ccc|ccc|ccc}
  \toprule
  \multicolumn{2}{c}{\multirow{2}{*}{\diagbox{Method}{Dataset}}} & \multicolumn{3}{c}{TACRED} & \multicolumn{3}{c}{TACRED-Revisit} & \multicolumn{3}{c}{Re-TACRED} & \multicolumn{3}{c}{SemEval}\\
  \cmidrule(r){3-5}\cmidrule(r){6-8}\cmidrule(r){9-11}\cmidrule(r){12-14}
  \multicolumn{2}{c}{} & K=2 & K=4 & K=8 & K=2 & K=4 & K=8 & K=2 & K=4 & K=8 & K=2 & K=4 & K=8 \\
  \midrule
  \multirow{2}{*}{{\centering\textbf{Mix 4}}}
  & Random & 14.65 & 18.49 & 27.25 & 16.49 & 16.83 & 25.81 & 18.01 & 30.77 & 37.28 & 29.12 & 37.99 & 36.58\\  
  & Ours & \textbf{14.95} & \textbf{18.73} & \textbf{30.46} & \textbf{19.72} & \textbf{20.85} & \textbf{27.36} & \textbf{20.89} & \textbf{34.73} & \textbf{39.34} & \textbf{32.97} & \textbf{43.74} & \textbf{66.98}  \\ 
  \midrule
  \multirow{2}{*}{{\centering\textbf{Mix P}}} 
  & Random & 19.08 & 23.06 & 26.01 & 21.43 & 21.77 & 27.11 & 22.65 & 26.63 & 42.95 & 21.66 & 23.87 & 29.93 \\  
  & Ours & \textbf{20.32} & \textbf{25.36} & \textbf{30.13} & \textbf{24.52} & \textbf{25.59} & \textbf{28.96} & \textbf{27.48} & \textbf{35.07} & \textbf{53.25} & \textbf{37.81} & \textbf{44.71} & \textbf{68.13}  \\ 
  \bottomrule
\end{tabular}}
\caption{\label{ablation}
Micro F1 (\%) of different algorithm of M-BRe framework on KnowPrompt. \textbf{Random} and \textbf{Ours} mean random and Algorithm \ref{alg:relation_grouping}-based Relation Grouping. \textbf{Mix} means hybrid training samples combining constructed and manually annotated samples, where P = $\lfloor N/6 \rfloor$ and $N$ is the total number of relation categories for each RE dataset.
}
\end{table*}

\begin{table}[t] 
\centering
\renewcommand{\arraystretch}{0.95}
\scalebox{0.83}{
\begin{tabular}{@{}lc|cccc@{}} 
\toprule
\multicolumn{2}{c}{Method}         & Tacred           & Tacrev           & Retacred          & SemEval      \\
\midrule
\multirow{2}{*}{{\centering\textbf{Pure 4}}}
  & Random & 7.71 & 9.25 & 8.38 & 17.05\\  
  & Ours & \textbf{9.80} & \textbf{11.61} & \textbf{14.03} & \textbf{19.15}\\
\midrule 
\multirow{2}{*}{{\centering\textbf{Pure P}}}
  & Random & 12.21 & 12.66 & 16.39 & 8.63\\  
  & Ours & \textbf{17.52} & \textbf{19.03} & \textbf{20.36} & \textbf{19.70}\\
\bottomrule
\end{tabular}}
\caption{\label{tab:ablation2}Micro F1 (\%) of different algorithm of M-BRe framework on KnowPrompt. \textbf{Pure} means purely constructed training samples.}
\end{table}

\paragraph{For Generated Training Samples.} Considering that the quality of the generated training samples is difficult to directly assess, we put the generated training samples into KnowPrompt \citep{DBLP:conf/www/ChenZXDYTHSC22} and RetrievalRE \citep{DBLP:conf/sigir/ChenLZTHSC22} for training. KnowPrompt achieves satisfying performance through Knowledge Injection and Synergistic Optimization, while RetrievalRE improves the model's generalization ability by combining retrieval enhancement and prompt tuning. Their performance on the test dataset reflects the quality of the framework-generated training samples. The better the performance of them, the higher the quality of the framework-generated training samples. Finally, we follow the existing RE studies and adopt Micro F1 as the main evaluation metric.

Meanwhile, in order to further evaluate the quality of the relation extraction training samples generated by the M-BRe framework, we reformat them into Supervised Fine-Tuning (SFT) datasets for multi-class classification to fine-tune LLMs. The observed improvement in the LLMs' multi-class classification performance post-SFT confirms the effectiveness of the generated samples.

\section{Results and Analysis}
\subsection{Main Results}
\label{sec:Main Results}
For M-BRe framework, we adopt 4-grouping and $\lfloor N/6 \rfloor$-grouping strategies to extract the relation categories from unlabeled sentence-level texts, which are then used to construct RE training samples. The sample sizes of RE datasets constructed by each LLM on identical unlabeled sentence-level texts are summarized in Appendix \ref{appendix:constructed_datasets}. The main results are detailed in Table \ref{main-result} and Table \ref{tab:pure}.
\paragraph{Comparing to Manually Labeled Training Samples.} In Table \ref{tab:pure}, Pure 4 and Pure P mean only using framework-constructed samples for training, and they differ in the number of groupings in ``Multi-Part'', while Knowprompt and RetrievalRE mean only using manually labeled training samples in Table \ref{main-result}. K denotes the number of manually labeled training samples for each relation category. The performance of Pure 4 and Pure P on all datasets can only match or exceed that of KnowPrompt and RetrievalRE when K = 2 or 4. We first analyze that the unlabeled sentence-level texts have not undergone a thorough data cleaning and screening process. The second point is that the distribution of relation categories extracted by the M-BRe framework is highly imbalanced, exhibiting a long-tail issue.

\paragraph{Comparing to Mixed Training Samples.} We mix manually labeled training samples with constructed training samples and use them to train RE models. This corresponds to Mix 4 and Mix P in Table \ref{main-result}. We observe that RE models' performance is higher than when only pure manually labeled training samples or constructed training samples are used under all settings. The results indicate that incorporating the constructed training samples with existing manually labeled training samples can strongly enhance the performance of RE models. However, the efficacy of the constructed training samples diminishes as the volume of manually labeled training data increase. Our analysis suggests that synthetically constructed training samples can effectively improve RE models' understanding of relation categories when manually labeled training data are scarce. As the volume of manually labeled data increase, the quality variability of the generated training samples introduce noise, due to uncleaned text sources. This results in potential misinterpretation of relation categories by RE models, ultimately leading to a slow performance improvement when using mixed training data.

\begin{table*}[t]

\centering
\renewcommand{\arraystretch}{1}
\scalebox{0.86}{
\begin{tabular}{ccccc|cccc}
  \toprule
  \multirow{2}*{Method} & \multicolumn{4}{c}{Qwen2.5-7B-Instruct-1M} & \multicolumn{4}{c}{Qwen2.5-14B-Instruct-1M} \\
  \cmidrule(r){2-5}\cmidrule(r){6-9}
  & Macro-P & Macro-R & Macro-F1 & \multicolumn{1}{c|}{Micro-F1}
  & Macro-P & Macro-R & Macro-F1 & Micro-F1 \\
  \cmidrule(r){1-9}
  \multicolumn{1}{c|}{Base} & 41.54 & 27.94 & 24.64 & 24.37 & 57.00 & 53.25 & 47.80 & 47.36\\
  \multicolumn{1}{c|}{Only Generated} & 50.72 & 47.64 & 45.32 & 43.95 & 61.84 & 57.37 & 58.22 & 56.22\\
  \multicolumn{1}{c|}{Only Manual} & 64.97 & 61.59 & 60.33 & 64.73 & 71.46 & 70.20 & 68.00 & 66.85\\
  \multicolumn{1}{c|}{Manual$\rightarrow$Generated} & 58.94 & 56.87 & 58.14 & 55.73 & 64.73 & 66.38 & 63.72 & 61.42\\
  \multicolumn{1}{c|}{Generated$\rightarrow$Manual} & \textbf{69.21} & \textbf{66.36} & \textbf{64.82} & \textbf{69.08} & \textbf{73.43} & \textbf{76.47} & \textbf{73.81} & \textbf{72.32}\\
  \bottomrule
\end{tabular}}
\caption{\label{SFT}
Micro F1 (\%) of each LLM with different SFT routes. \textbf{Base} means LLM without SFT. \textbf{Only Generated} means only using LLMs-generated training samples for SFT. \textbf{Only Manual} means only using manually labeled training samples for SFT. \textbf{Manual$\rightarrow$Generated} means using LLMs-generated training samples first and then manually labeled training samples for SFT. \textbf{Generated$\rightarrow$Manual} means using manually labeled training samples first and then LLMs-generated training samples for SFT.
}
\end{table*}

\paragraph{Comparing to Number of Groups.} We establish relation category groupings with sizes 4 and P, corresponding to Mix 4 and Mix P in Table \ref{main-result}, along with Pure 4 and Pure P in Table \ref{tab:pure}. The experimental results demonstrate that Pure P consistently outperforms Pure 4 across all datasets. We attribute this superiority to Pure P's larger number of groupings, which provides greater discriminability among relation categories, thereby enabling LLMs to make more accurate judgments. Moreover, under all few-shot settings, Mix P outperforms Mix 4, although the performance gap diminishes with increasing K. We attribute this to the reduced contribution of LLMs-generated training samples in RE model training as more manually labeled data become available, where certain inevitable noise even exerts negative effects on model training. This phenomenon demonstrate that the number of groupings significantly impacts RE performance of M-BRe framework. Therefore, we conduct a comprehensive experimental analysis into the effect of grouping quantity in ``§\ref{section:Number of Groupings} Number of Groupings''.

\subsection{Ablation Analysis}
\begin{figure}[t]
\centering
\centerline{\includegraphics[scale=0.2]{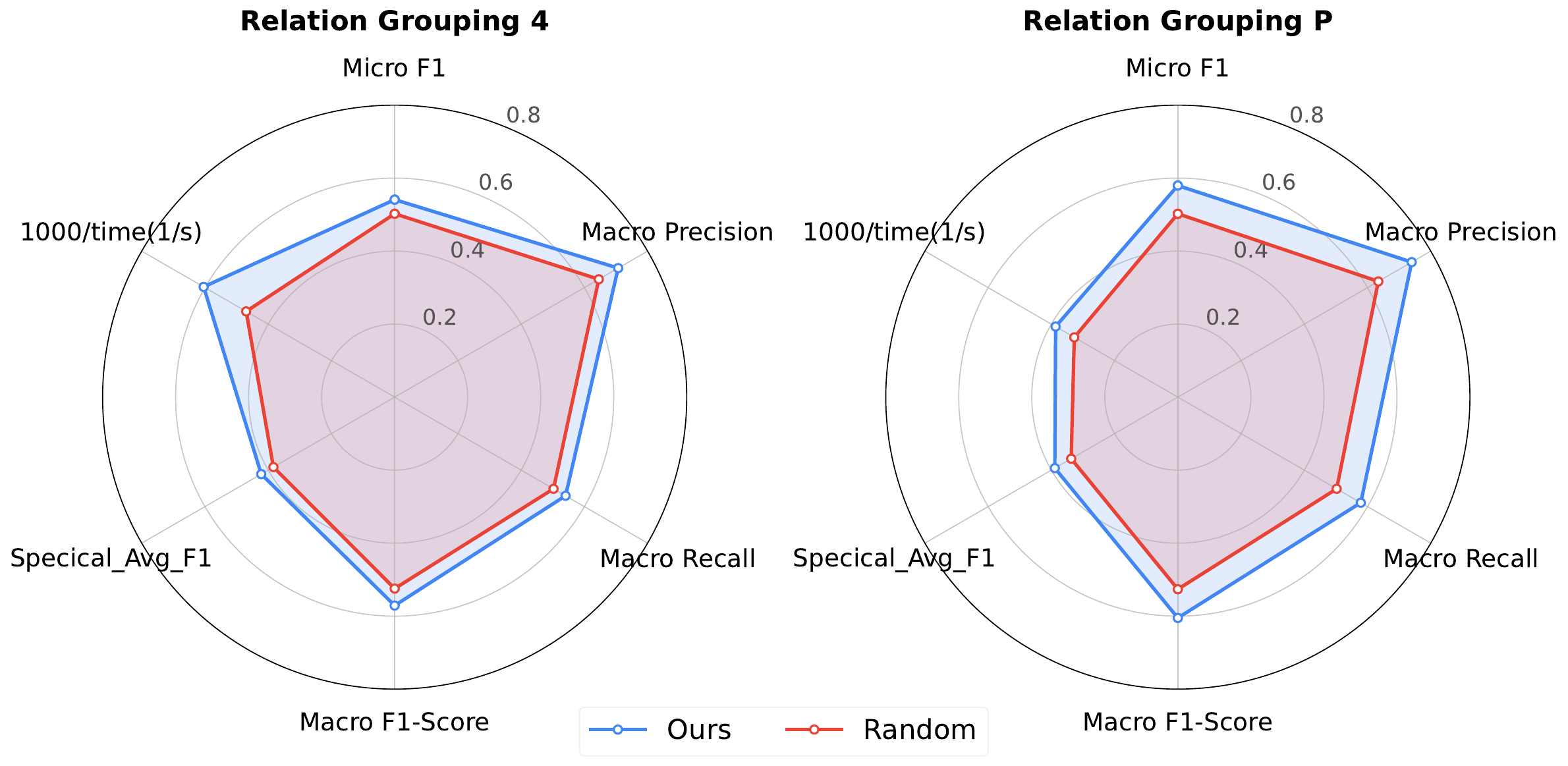}}
\caption{Comparative evaluation of the two methods' effects on M-BRe framework performance. \textbf{Ours} and \textbf{Random} represent Algorithm \ref{alg:relation_grouping}-based and random Relation Grouping in the M-BRe framework.}
\label{fig:ablation1}
\end{figure}

In order to verify the effectiveness of Relation Groupings Algorithm, we conduct comprehensive ablation experiments: relation categories were divided into 4 and P groups using both the Relation Grouping Algorithm and other grouping ways. As demonstrated in Figure \ref{fig:ablation1} and Appendix \ref{appendix:clustering}, our method significantly enhances the relation extraction capability of the M-BRe framework across all evaluation metrics, while simultaneously achieving faster processing speed than the random grouping approach. 

To further validate the effectiveness of our approach, we employed random-based M-BRe framework on the same unlabeled sentence-level texts to extract relation categories and construct RE training samples with Qwen2.5-7B-Instruct-1M. These samples were subsequently utilized to train KnowPrompt for comparative performance evaluation. As shown in Table \ref{ablation} and \ref{tab:ablation2}, Ours (Pure 4, P) consistently outperform Random (Pure 4, P), demonstrating that the constructed RE training samples with Algorithm \ref{alg:relation_grouping}-based Relation Grouping achieve higher quality than those with random Relation Grouping, under both grouping sizes of 4 and P. Ours (Mix 4, P) also consistently exhibit superior performance to Random (Mix 4, P) in all settings, further validating the aforementioned conclusions.

\begin{figure*}[t]
\centering
\begin{subfigure}[b]{0.325\textwidth} 
    \centering
    \includegraphics[width=\textwidth]{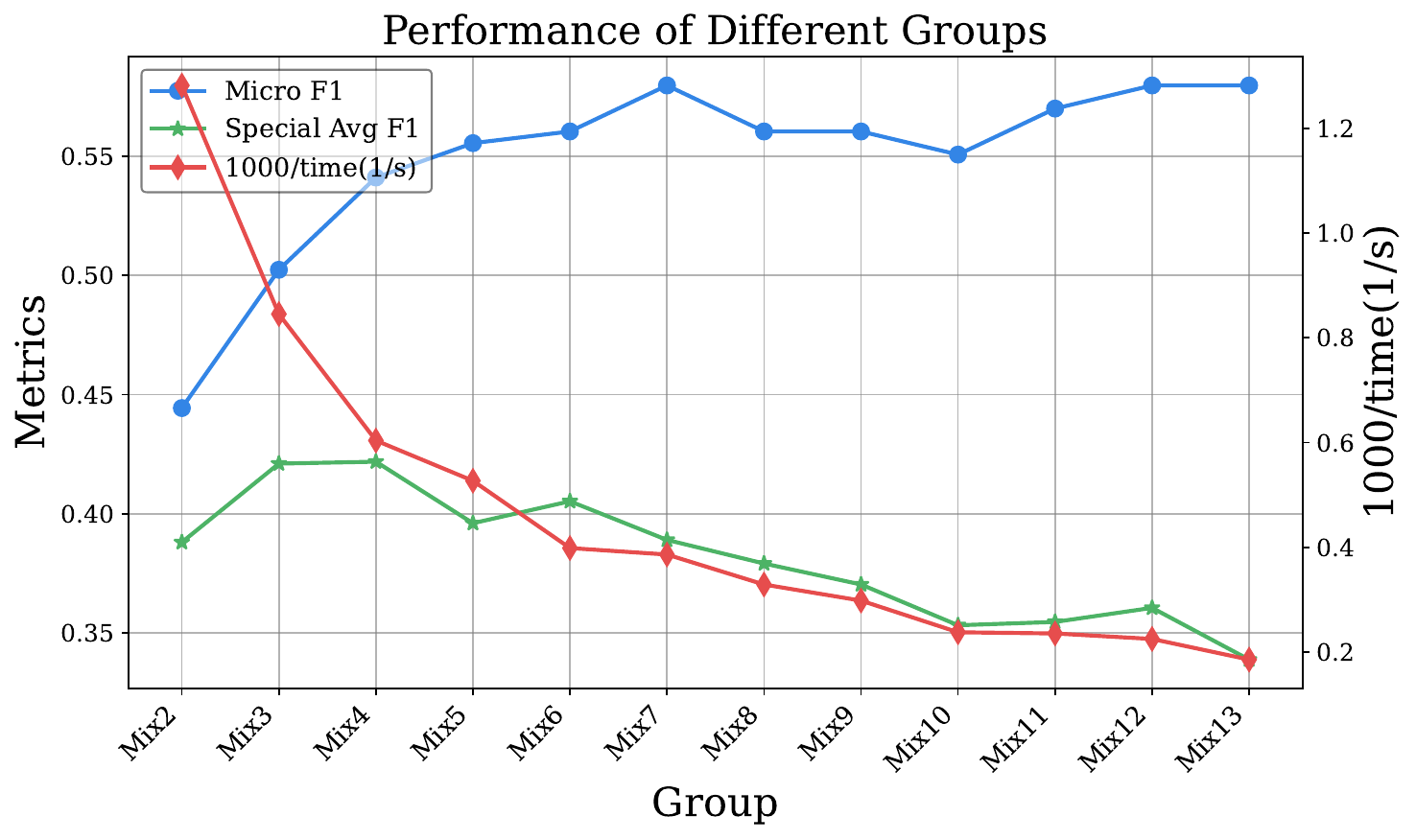}
    \caption{Qwen2.5-7B-Instruct-1M / Tacred}
    \label{fig:groups1}
\end{subfigure}
\hfill 
\begin{subfigure}[b]{0.325\textwidth}
    \centering
    \includegraphics[width=\textwidth]{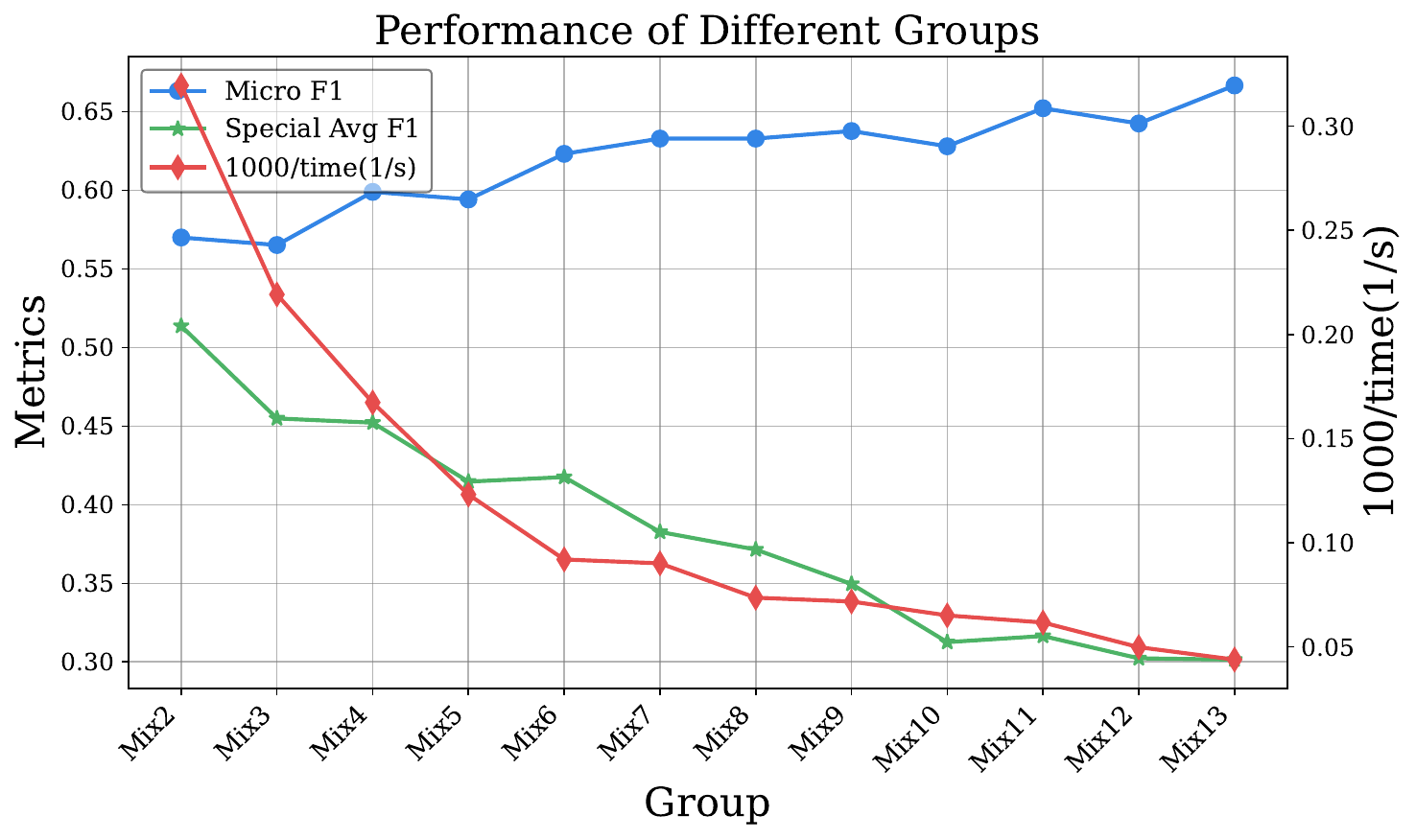}
    \caption{Qwen2.5-14B-Instruct-1M / Tacred}
    \label{fig:groups3}
\end{subfigure}
\hfill 
\begin{subfigure}[b]{0.325\textwidth}
    \centering
    \includegraphics[width=\textwidth]{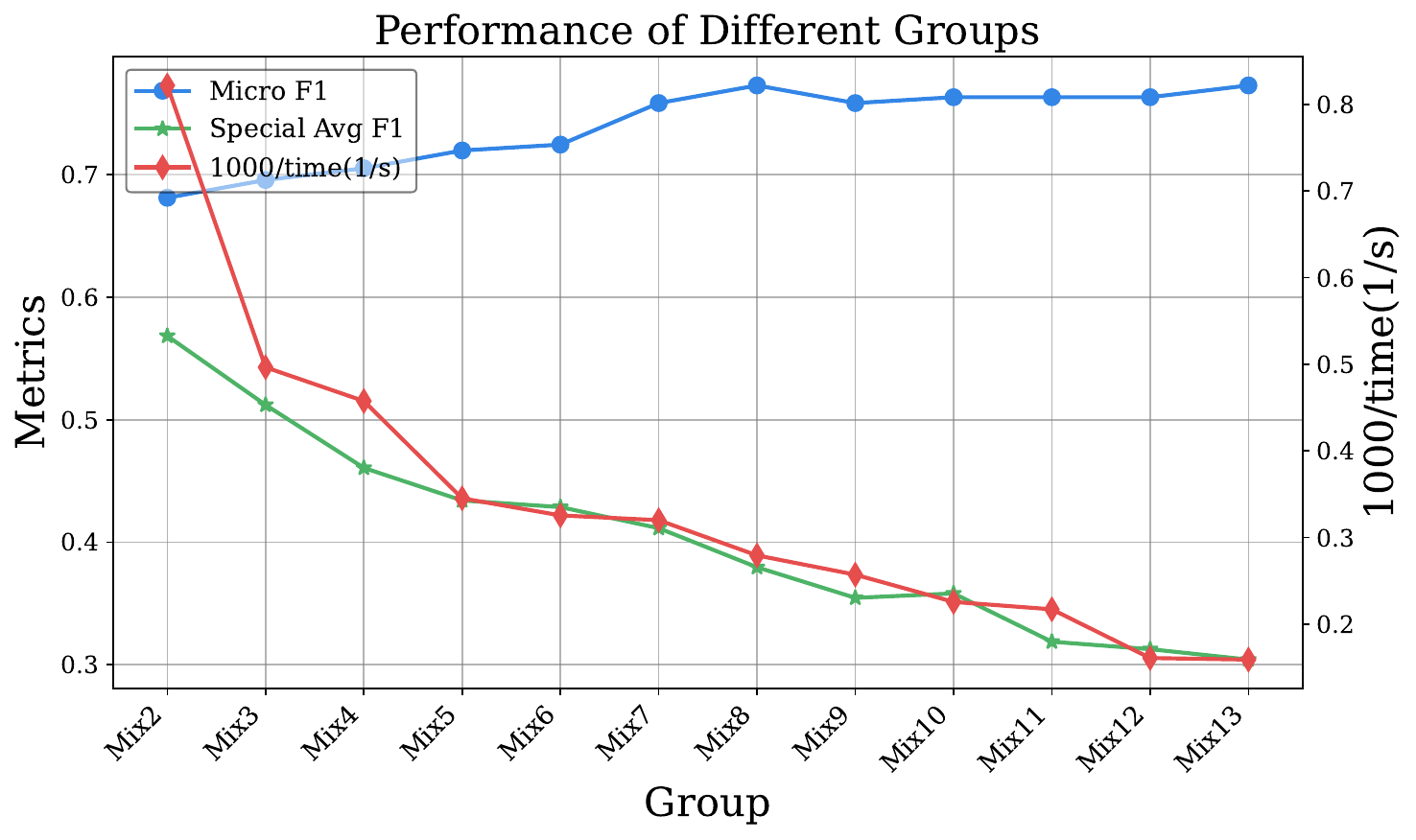}
    \caption{Qwen3-14B / Tacred}
    \label{fig:groups3}
\end{subfigure}
\hfill 
\begin{subfigure}[b]{0.325\textwidth}
    \centering
    \includegraphics[width=\textwidth]{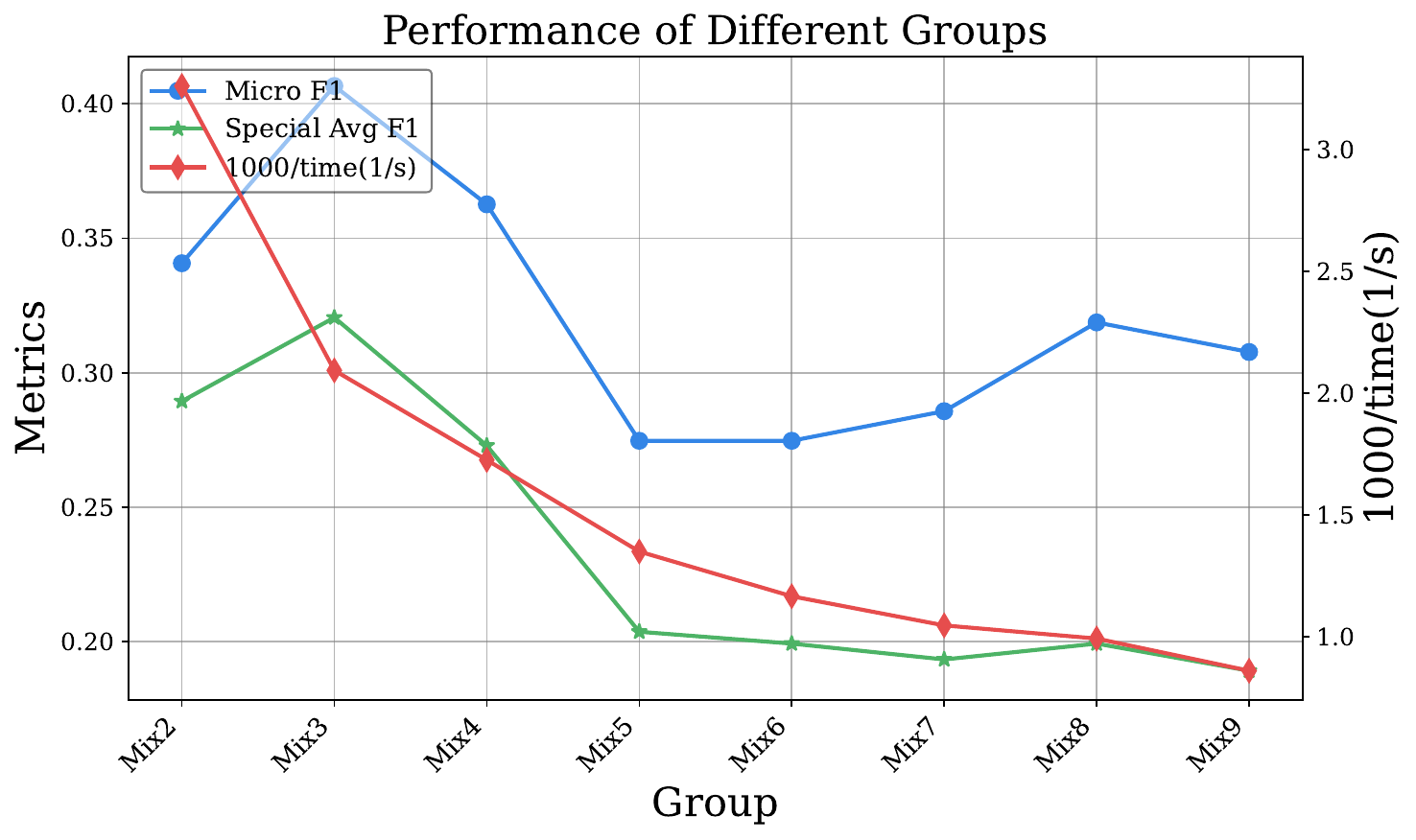}
    \caption{Qwen2.5-7B-Instruct-1M / SemEval}
    \label{fig:groups2}
\end{subfigure}
\hfill 
\begin{subfigure}[b]{0.325\textwidth}
    \centering
    \includegraphics[width=\textwidth]{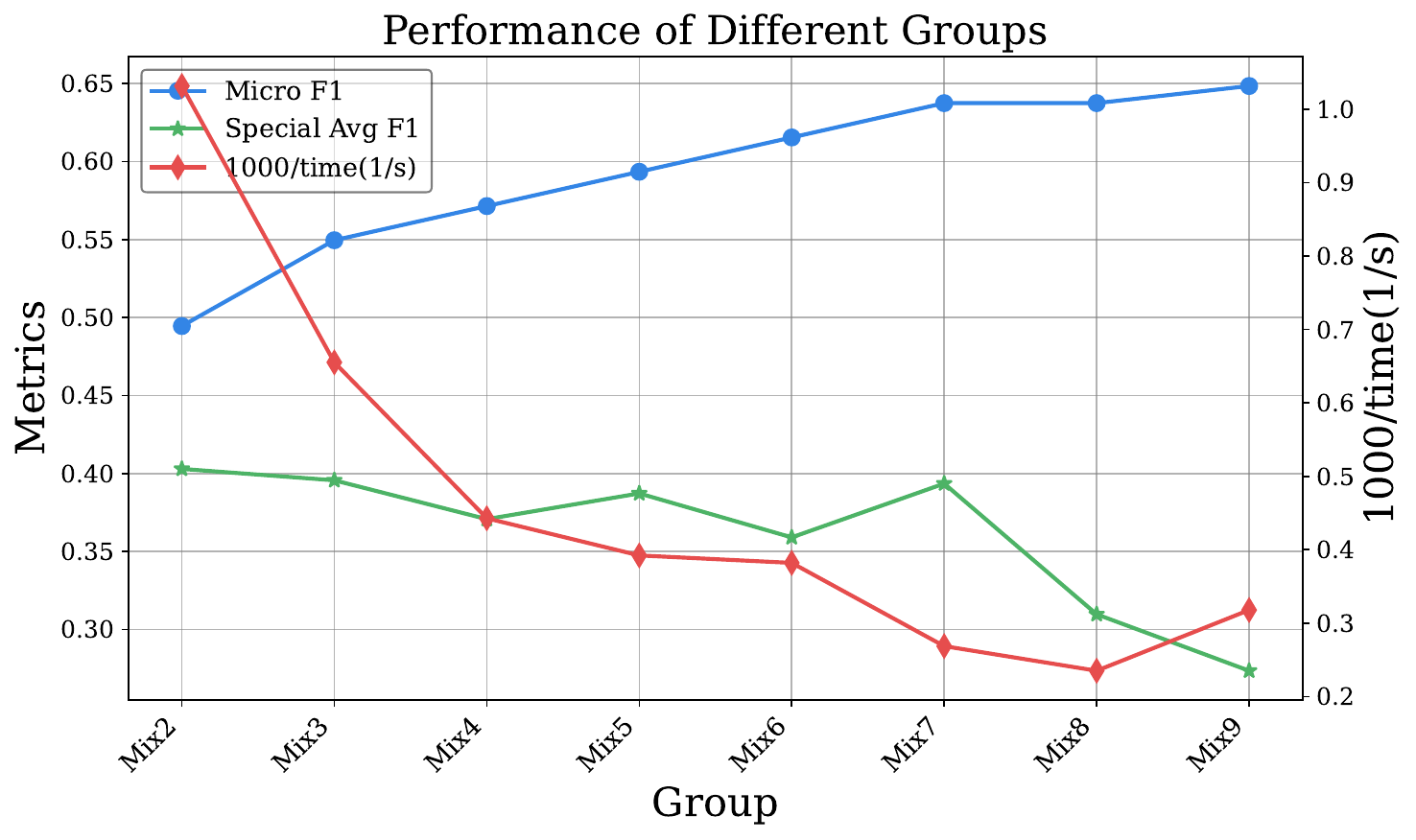}
    \caption{Qwen2.5-14B-Instruct-1M / SemEval}
    \label{fig:groups2}
\end{subfigure}
\hfill 
\begin{subfigure}[b]{0.325\textwidth}
    \centering
    \includegraphics[width=\textwidth]{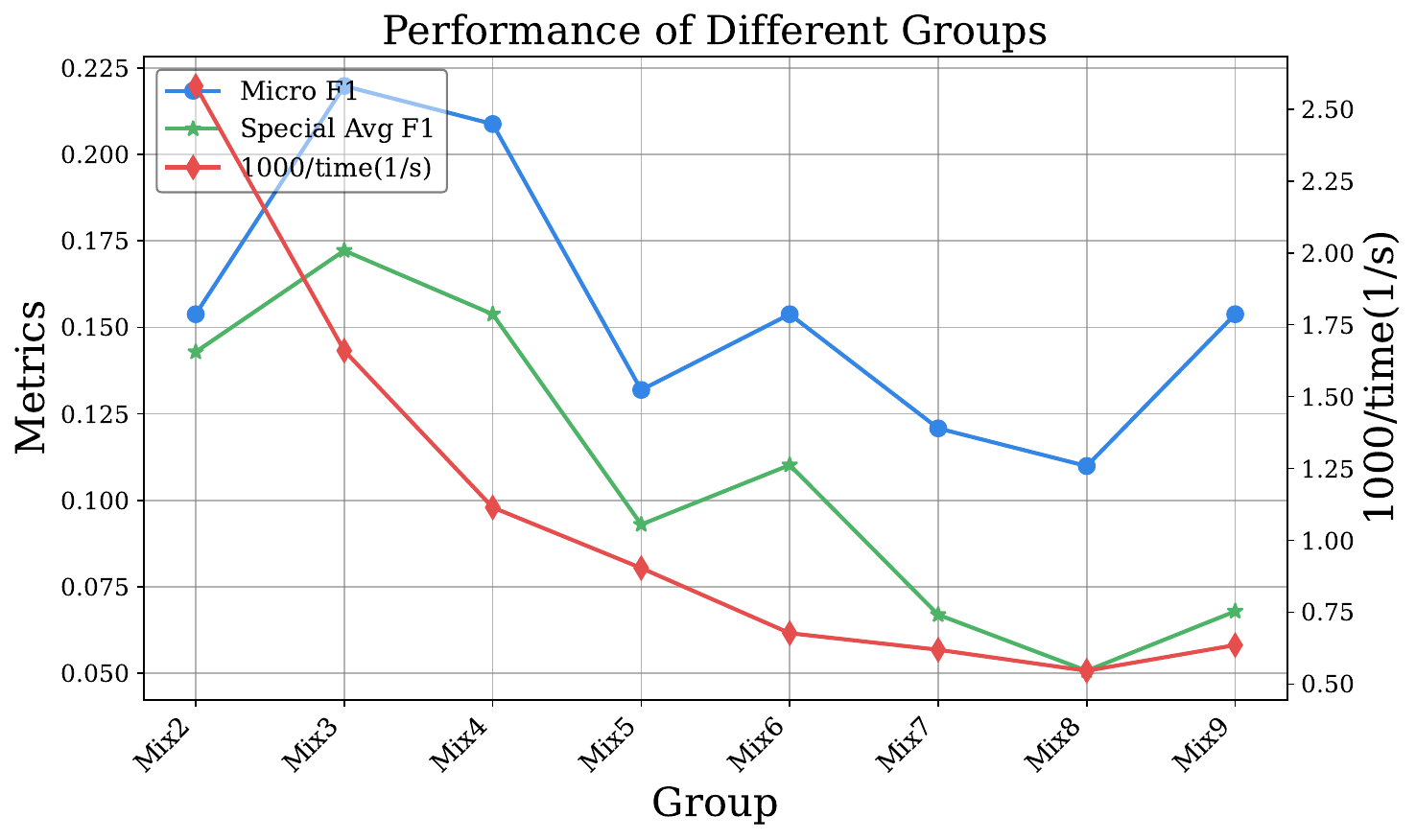}
    \caption{Qwen3-14B / SemEval}
    \label{fig:groups2}
\end{subfigure}
\hfill 
\caption{Micro F1 (\%) and Special\_Avg\_F1 (\%) of different number of groupings on each LLM.}
\label{fig:groups}
\end{figure*}

\subsection{Number of Groups}
\label{section:Number of Groupings}
Evidently, the number of relation category groups significantly impacts the overall performance of the M-BRe framework. Therefore, we conducted comprehensive experiments while controlling this variable as shown in Figure \ref{fig:groups} and Appendix \ref{appendix:Number of Groupings}. The relation extraction performance of the M-BRe framework gradually improves with increasing number of groups. At $\lfloor N/6 \rfloor$ groups, the performance essentially peaks, matching that of the binary classification approach while achieving more than twice the processing speed. Our analysis shows that at $\lfloor N/6 \rfloor$ groups, the M-BRe framework combines the advantages of both multi-class and binary classification, enabling rapid relation extraction from unlabeled sentence-level texts while maintaining accuracy. However, the framework's performance begins to gradually degrade as the number of groupings continues to increase. We attribute this to the lengthened output list in the Multi-prompt phase as group numbers increase, leading to accumulated error propagation. This phenomenon demonstrates the existence of an inflection point for group numbers, confirming that more groups do not necessarily yield better results.

\subsection{Cold Start for SFT}
We reconstruct the training samples generated by the M-BRe framework and the 8-shot manually labeled training samples into an SFT dataset, then fine-tune Qwen2.5-7B-Instruct-1M and Qwen2.5-14B-Instruct-1M through the following four routes:
\begin{itemize}[itemsep=0pt]
  \item Using only LLMs-generated training samples.
  \item Using only manually labeled training samples.
  \item Using LLMs-generated training samples first and then manually labeled training samples.
  \item Using manually labeled training samples first and then LLMs-generated training samples.
\end{itemize}
As shown in Table \ref{SFT}, all SFT routes significantly improve the performance of LLMs on direct multi-class relation extraction, further demonstrating the effectiveness of LLMs-generated training samples. Admittedly, the performance after SFT with manually labeled training samples surpasses that with LLMs-generated training samples, which we attribute to the two major characteristics of LLMs-generated training samples mentioned in ``§\ref{sec:Main Results} Main Results''. Meanwhile, it is encouraging to observe that the performance after two-stage fine-tuning (first with LLMs-generated then manually labeled samples) slightly outperforms using only manually labeled samples. We posit that the LLMs-generated training samples serve as a cold start for the second-stage fine-tuning, thereby effectively enhancing LLMs' performance.

\section{Conclusion}
In this paper, we propose a novel M-BRe framework for constructing RE training samples with LLMs. This framework effectively combines the advantages of multi-class and binary classification to efficiently utilize unlabeled texts to acquire sentence-level RE training samples, particularly through the Relation Grouping Module and the Label Decision Module. The former enables LLMs to rapidly comprehend and distinguish different relations while making an accurate classification judgment. The latter allows LLMs to verify their judgments while accommodating cases where single head-tail entity pair may correspond to multiple valid relations. Experimental results prove the effectiveness of RE training samples constructed by the M-BRe framework in few-shot scenarios.

In future work, we plan to explore the following directions: (1) more rational and effective approaches for Relation Grouping; (2) alternative methods for relation category judgment that outperform Confidence-based approaches.
\section*{Limitations}
Despite our best efforts, the proposed M-BRe framework in this paper still has several limitations.

\textbf{Long-Tail Issue:} Constrained by the nature of unlabeled sentence-level texts, the distribution of RE training samples constructed by B-MRe framework exhibits a long-tail issue, where the scarcity of instances for certain relation categories may limit the performance of RE models.

\textbf{LLMs:} Although we have enabled the M-BRe framework to efficiently construct RE training samples, the quality of these constructed samples remains significantly influenced by the inherent capabilities of the open-source LLMs themselves.



\section*{Acknowledgements}
This research is supported by the National Natural Science Foundation of China (No. 62306140, No. 62476127), the Natural Science Foundation of Jiangsu Province (No. BK20242039), the Basic Research Program of the Bureau of Science and Technology (ILF24001), the Fundamental Research Funds for the Central Universities (No. NJ2023032), the Scientific Research Starting Foundation of Nanjing University of Aeronautics and Astronautics (No. YQR21022), the Key Project of Jiangsu Collaborative Innovation Center of Chinese Medicinal Resources Industrialization (No. 000003401025-6), the Open Project of Chinese Materia Medica First-Class Discipline of Nanjing University of Chinese Medicine (No. ZYXJC2024-010) and the High Performance Computing Platform of Nanjing University of Aeronautics and Astronautics.

\bibliography{anthology,custom}
\bibliographystyle{acl_natbib}

\appendix
\section{Experimental Details}
\subsection{Datasets}
\label{appendix:datasets}
For comprehensive experiments, we conducted experiments on four relation extraction datasets: TACRED, TACRED-Revisit, Re-TACRED and SemEval 2010 Task 8 (SemEval). The statistics of the RE datasets are shown in Table \ref{tab:dataset}. A brief introduction to these data is given below:

\paragraph{TACRED:} a large-scale sentence-level relation extraction dataset from the annual TACBP4 challenge, containing over 106,000 sentences. It involves 42 different relation categories, including 41 common relation categories and a special ``no relation'' relation category.

\paragraph{TACRED-Revisit:} a dataset constructed on the basis of the TACRED dataset. The researchers found errors in the development and test sets of the original TACRED dataset and corrected them while keeping the training set intact.

\paragraph{Re-TACRED:} another version of the TACRED dataset, which addresses some of the shortcomings of the original TACRED dataset by reconstructing the training, validation and test sets. Meanwhile, this dataset removes the original 6 relation categories and adds 4 new relation categories to the TACRED dataset, so that a dataset with 40 relation categories is finally obtained.

\paragraph{SemEval:} a traditional relation extraction dataset, containing 10,717 annotated samples, covers 9 bi-directional relation categories and a special ``no relation'' relation category.
\begin{table}[h]

\centering
\renewcommand{\arraystretch}{1}
\scalebox{0.8}{
\begin{tabular}{lrrrc}
\hline
Dataset & Train& Val & Test &Relation\\
\hline
SemEval & {6,507}& {1,493}& {2,717}& {19} \\ 
TACRED & {68,124} & {22,631}& {15,509}& {42}\\
TACRED-Revisit & {68,124}& {22,631}& {15,509}& {42} \\
Re-TACRED & {58,465}& {19,584}& {13,418}& {40} \\ 
\hline
\end{tabular}}
\caption{\label{tab:dataset}Statistics of the RE datasets. Including the numbers of instances in different splits and the numbers of relation categories.}

\end{table}

\subsection{Implementation Details}
\label{appendix:Implementation Details}
\paragraph{For KnowPrompt and RetrievalRE.} We follow \citep{DBLP:conf/www/ChenZXDYTHSC22,DBLP:conf/sigir/ChenLZTHSC22} and use RoBERTA\_LARGE \citep{DBLP:journals/corr/abs-1907-11692} in all experiments for comparison. 

\paragraph{For Large Language Models.} Considering the cost requirements and the strength of the different LLMs themselves, we used Qwen2.5-7B-Instruct-1M, Qwen2.5-14B-Instruct-1M \citep{DBLP:journals/corr/abs-2501-15383}, Qwen3-14B in our experiments, setting temperature = 0.6. For post-processing, since the unlabeled text has not been filtered, the following two data distributions appear after extracting relation categories with the B-MRe framework: (1) Most sentence-level texts are meaningless, so the number ratio of ``NA'' to ``non-NA'' is relatively large. (2) In sentence-level text, the frequency of different ``non-NA'' relation categories is different, so their number shows a long-tailed distribution. To combat this, we calculated the average number of ``non-NA'' relation categories. This number of samples are randomly selected from the ``NA'' relation category and combined with the ``non-NA'' relation category to form the final training data.

\paragraph{For Supervised Fine-Tuning.} All models are fine-tuned using the LLaMA-Factory framework \cite{llamafactory} on 8 NVIDIA RTX 3090 GPUs. We employ Low-Rank Adaptation (LoRA) \cite{LoRA} with rank \(r=16\) applied to all linear layers. Other hyperparameters include a batch size of 4, a learning rate of \( 10^{-4} \), and 3 training epochs.

\subsection{Statistics on the number of samples constructed by each LLM}
\label{appendix:constructed_datasets}
The number of unlabeled sentences used in our experiments is 4401. Tabel \ref{tab:generated} details the number of generated RE training samples for each LLM, it can be seen that the existence of relations within unlabeled sentences is indeed very sparse. Table \ref{tab:constructed} details the number of constructed RE training samples for each LLM. Our analysis reveals that individual LLMs with stronger performance exhibit enhanced capability in constructing RE training samples from unlabeled text, whereas the grouping factor demonstrates no statistically significant impact on the yield of RE training instances.

\begin{table*}[ht] 

\centering
\renewcommand{\arraystretch}{1.2}
\scalebox{0.9}{
\begin{tabular}{@{}lcccc@{}} 
\toprule
LLMs         & Dataset           & Number of non-NA	&Relevant ratio	&Relevant quantity      \\
\midrule
\multirow{2}{*}{{\parbox{2.28cm}{\textbf{Qwen2.5-7B-Instruct-1M}}}} &
Tacred 4/P & 807/799&	151/146	&5/6\\
& Semeval 4/P & 628/498	&152/144	&3/3\\
\midrule 
\multirow{2}{*}{{\parbox{2.28cm}{\textbf{Qwen2.5-14B-Instruct-1M}}}} &
 Tacred 4/P & 	855/974	&102/253&	3/2\\
 & Semeval 4/P & 1561/1657	&403/224	&0/0\\   
\midrule 
\multirow{2}{*}{{\parbox{2.28cm}{\textbf{Qwen3-14B}}}} &
 Tacred 4/P & 890/827&	122/110&	6/6\\
 & Semeval 4/P & 803/1041	&330/197	&5/4\\
\bottomrule
\end{tabular}}
\caption{\label{tab:generated} \textbf{Number of non-NA} represents the number of non-NA relation categories, \textbf{relevant ratio} represents the number ratio of relations with maximum generated quantities to relations with minimum generated quantities, \textbf{relevant quantity} represents the number of relations generated with a quantity of 0.
}
\end{table*}

\begin{table}[ht] 

\centering
\renewcommand{\arraystretch}{0.95}
\scalebox{0.86}{
\begin{tabular}{@{}lcccc@{}} 
\toprule
LLMs         & Tacred           & Tacrev           & Retacred          & SemEval      \\
\midrule

\multirow{2}{*}{{\parbox{2.28cm}{\textbf{Qwen2.5-7B-Instruct-1M}}}} &
411 & 411 & 376 & 670\\
& 429 & 429 & 383 & 531\\
\midrule 
\multirow{2}{*}{{\parbox{2.28cm}{\textbf{Qwen2.5-14B-Instruct-1M}}}} &
 476 & 476 & 448 & 1648\\
 & 510 & 510 & 445 & 1749\\   
\midrule 
\multirow{2}{*}{{\parbox{2.28cm}{\textbf{Qwen3-14B}}}} &
 915 & 915 & 833 & 865\\
 & 851 & 851 & 759 & 1115\\
\bottomrule
\end{tabular}}

\caption{\label{tab:constructed}Number of samples constructed by each LLM. The first and second rows of each LLM correspond to group 4 and group p, where p $=
\lfloor N/6 \rfloor$ and $N$ is the total number of relation categories for each RE dataset.}

\end{table}

\subsection{Relation Explanation}
\label{sub:RES}
We give explanations for each relation in the four datasets. The detailed explanation for each relation is shown in Table \ref{tab:Relation Explanation}.

\section{Experimental Results}
\subsection{Number of Groupings}
\label{appendix:Number of Groupings}
Table \ref{tab:multi-binary} details the concrete relation extraction performance of the M-BRE framework, evaluating the comprehensive metrics under different number of groupings. Regarding the Tacred and SemEval datasets, our experimental configurations employ group sizes of [2, 13] and [2, 9] respectively.

\begin{table*}[t]

\centering
\renewcommand{\arraystretch}{1.2}
\scalebox{0.9}{
\begin{tabular}{cccc|ccc}
  \toprule
  \multirow{2}*{Method} & \multicolumn{3}{c}{Qwen2.5-7B-Instruct-1M for Tacred} & \multicolumn{3}{c}{Qwen2.5-7B-Instruct-1M for SemEval} \\
  \cmidrule(r){2-4}\cmidrule(r){5-7}
  & Mi-F1 & Ma-F1 & \multicolumn{1}{c|}{S\_A\_F1}
  & Mi-F1 & Ma-F1 & S\_A\_F1 \\
  \cmidrule(r){1-7}
  \multicolumn{1}{c|}{K-Means 4/P} &48.79/51.21	 &50.65/53.02 &	40.71/37.80	 &20.88/37.36 &	18.38/31.64	 &19.78/31.07\\
  \multicolumn{1}{c|}{Hierarchical 4/P} & 48.31/50.72	&50.72/52.62	&40.39/38.58	&28.57/26.37	&23.97/23.21	&26.37/25.64\\
  \multicolumn{1}{c|}{Ours 4/P} & \textbf{57.10/60.50}&	\textbf{54.05/57.86}	&\textbf{42.19/38.91}	&\textbf{30.98/39.52}&	\textbf{34.74/38.95}	&\textbf{27.29/32.05}\\
  \cmidrule(r){1-7}
  \multirow{2}*{Method} & \multicolumn{3}{c}{Qwen2.5-14B-Instruct-1M for Tacred} & \multicolumn{3}{c}{Qwen2.5-14B-Instruct-1M for SemEval} \\
  \cmidrule(r){2-4}\cmidrule(r){5-7}
  & Mi-F1 & Ma-F1 & \multicolumn{1}{c|}{S\_A\_F1}
  & Mi-F1 & Ma-F1 & S\_A\_F1 \\
  \cmidrule(r){1-7}
  \multicolumn{1}{c|}{K-Means 4/P} & 54.11/57.97	&58.30/62.08	&42.09/35.36	&47.25/51.65	&44.05/47.72	&30.29/32.86\\
  \multicolumn{1}{c|}{Hierarchical 4/P} & 53.62/58.94&	57.79/62.71	&42.24/38.23	&50.04/46.15	&51.61/41.98	&33.92/35.77\\
  \multicolumn{1}{c|}{Ours 4/P} & \textbf{61.35/65.55}&\textbf{59.76/63.10}&	\textbf{45.22/38.27}	&\textbf{51.12/52.75}	&\textbf{54.74/56.84}	&\textbf{37.07/39.56} \\
  \cmidrule(r){1-7}
  \multirow{2}*{Method} & \multicolumn{3}{c}{Qwen3-14B for Tacred} & \multicolumn{3}{c}{Qwen3-14B for SemEval} \\
  \cmidrule(r){2-4}\cmidrule(r){5-7}
  & Mi-F1 & Ma-F1 & \multicolumn{1}{c|}{S\_A\_F1}
  & Mi-F1 & Ma-F1 & S\_A\_F1 \\
  \cmidrule(r){1-7}
  \multicolumn{1}{c|}{K-Means 4/P} & 67.63/71.01&	70.29/73.68	&42.46/40.90&	19.57/20.88&	17.79/18.74	&14.65/16.21\\
  \multicolumn{1}{c|}{Hierarchical 4/P} & 64.73/70.53&	68.08/74.66	&42.82/40.92	&18.27/20.47	&21.58/22.49	&\textbf{19.01/20.88}\\
  \multicolumn{1}{c|}{Ours 4/P} & \textbf{70.68/76.89}&	\textbf{70.95/76.19}	&\textbf{46.07/41.15}&	\textbf{19.76/21.16}&	\textbf{24.21/25.26}&	15.38/17.22 \\
  \bottomrule
\end{tabular}}
\caption{\label{tab:clusting}
Performance Results of Different Relation Grouping Methods for group 4 and group p, where p $=
\lfloor N/6 \rfloor$ and $N$ is the total number of relation categories for each RE dataset. Mi-F1, Ma-F1 and S\_A\_F1 represent Micro F1, Macro F1 and Special\_Avg\_F1.
}
\end{table*}

\subsection{Case Study}
\label{appendix:case study}
In response to the special phenomenon of ``Multi-Yes'' mentioned in §\ref{section:Confidence Judgment Module} Confidence Judgement Module, we conduct a case study on RE training samples constructed by the M-BRe framework.

\begin{figure*}[t]
\centering
\centerline{\includegraphics[scale=0.42]{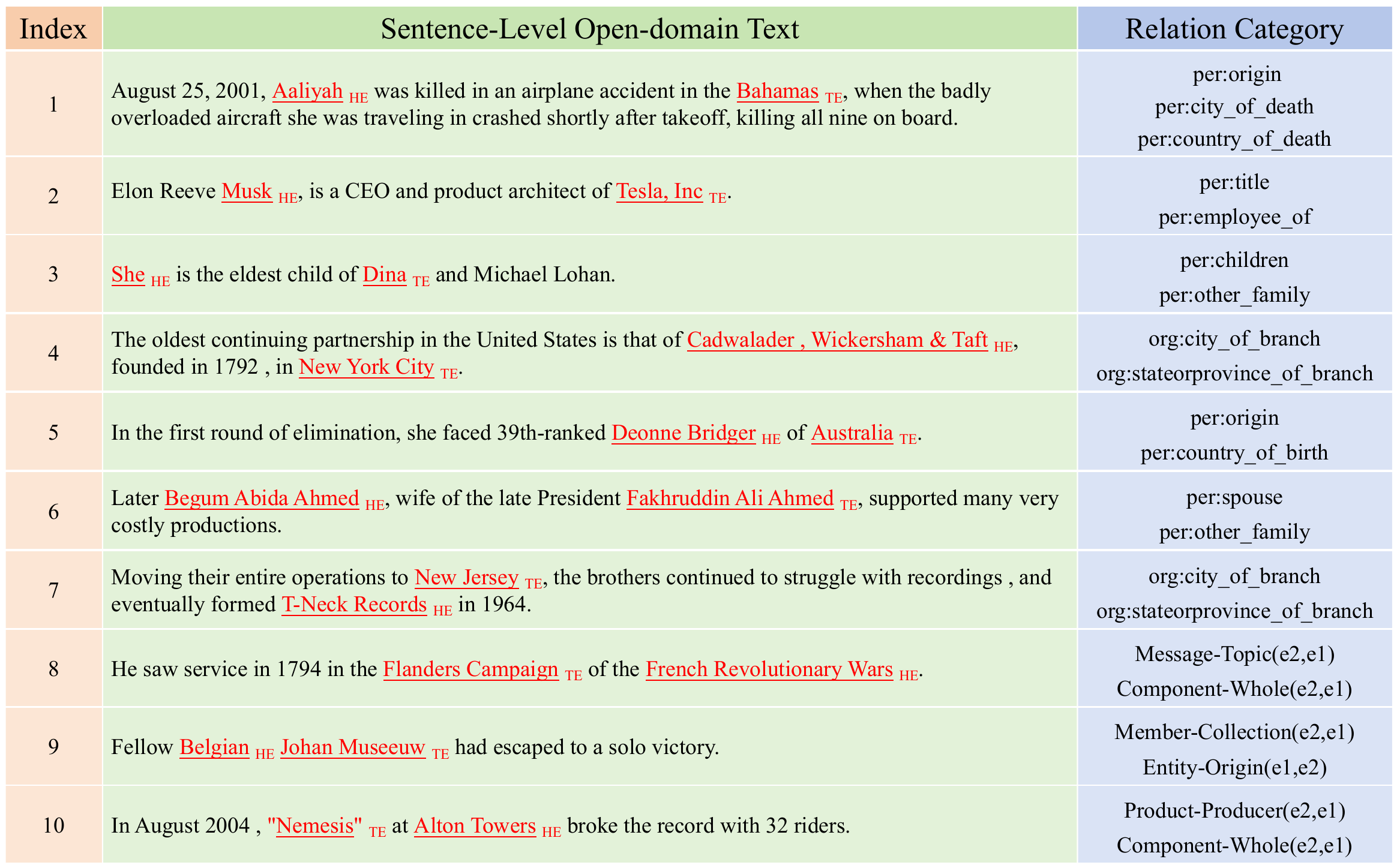}}
\caption{A case study of relation extraction on sentence-level unlabeled text with the M-BRe framework, where \textcolor{red}{HE} and \textcolor{red}{TE} represents the head entity and the tail entity.}
\label{fig:case study}
\end{figure*}

As shown in Figure \ref{fig:case study}, some head-tail entity pairs from unlabeled texts can be plausibly explained by multiple relation categories. Our M-BRe framework does not restrict the LLMs to output single relation category or ``NA/no\_relation/Others''. By granting the LLMs greater creative diversity, more and more valid high-quality RE training samples can be constructed from single sentence-level unlabeled text.





\subsection{Other Grouping Methods}
\label{appendix:clustering}
We have supplemented experiments on the relation grouping algorithms, including K-Means and hierarchical clustering. The experiments were conducted with the number of clusters set to 4 and $\lfloor N/6 \rfloor$, using Qwen2.5-7B-Instruct-1M, Qwen2.5-14B-Instruct-1M, and Qwen3-14B. The primary evaluation metrics were Micro-F1, Macro-F1, and Special\_Avg\_F1. As shown in Table \ref{tab:clusting}. Our relation grouping algorithm outperforms both K-Means and hierarchical clustering across most experimental conditions. This further validates the superiority of our proposed approach.

\section{Relation Grouping Algorithm}
\label{algorithm:RG}
The detailed Relation Grouping Algorithm is given in Algorithm \ref{alg:relation_grouping}.

\begin{algorithm}[h]
\caption{Relation Grouping}
\label{alg:relation_grouping}
\textbf{Input:} Relation list $R$, $k \gets \lfloor |R|/6 \rfloor$\\
\textbf{Output:} Groups $G = \{g_1,...,g_k\}$

\begin{algorithmic}[1]
\STATE $V \gets \text{TF-IDF}(R)$ \COMMENT{Vectorize relations}
\STATE $S \gets \text{cosine}(V)$, \\$D \gets 1-S$ \COMMENT{Similarity matrices}

\STATE $G \gets \{\varnothing\}^k$ \COMMENT{Initialize groups}
\STATE $(i,j) \gets \arg\max(D)$ \COMMENT{Select seeds}
\STATE $g_1 \gets \{R_i\}, g_2 \gets \{R_j\}$

\WHILE{$\exists$ unassigned relations}
\FOR{$r_u \in \text{unassigned}$}
\STATE $g^* \gets \arg\min_{g_i} \max_{r_g \in g_i} S[r_u][r_g]$
\ENDFOR
\STATE $g^* \gets g^* \cup \{r^*\}$ \COMMENT{Assign best candidate}
\ENDWHILE

\STATE $G \gets \text{sort}(R[\text{indices}])$\\ \COMMENT{Recover original relations}
\RETURN $G$
\end{algorithmic}
\end{algorithm}

\section{Resource Consumption}
\label{resource}
The detailed Resource Consumption is given in Table \ref{tab:resource consumption}.

\section{M-BRe for Fine-Grained Named Entity Recognition}
We applied the M-BRe framework (Qwen2.5-14B-Instruct-1M) to Fine-Grained Named Entity Recognition. Experimental results on two datasets are shown in Table \ref{tab:fgner}. It can be observed that the M-BRe framework also demonstrates high performance with low computational and time cost.

\begin{table}[ht] 
\centering
\renewcommand{\arraystretch}{1.1}
\scalebox{0.72}{
\begin{tabular}{@{}lcccccc@{}} 
\toprule
Datasets         & Methods & Precison           & Recall           & S\_A\_F1        & Time(h)      \\
\midrule

\multirow{3}{*}{{\parbox{2.28cm}{\textbf{BNN}}}} &
Multi & 28.10 & 1.27 & 2.39 & 0.14\\
&M-BRe & 30.20 & 40.00 & 32.04 & 2.52\\
&Binary & 26.09 & 94.05 & 37.93 & 5.18\\
\midrule 
\multirow{3}{*}{{\parbox{2.28cm}{\textbf{OntoNotes}}}} &
Multi & 31.48 & 0.70& 1.36& 0.22 \\
 &M-BRe& 30.91 & 23.81 & 24.03& 4.12\\ 
 &Binary& 27.97 & 79.80 & 37.73& 7.55 \\ 
\bottomrule
\end{tabular}}
\caption{\label{tab:fgner}Performance comparison of M-BRe Framework on Fine-Grained Named Entity Recognition.}
\end{table}

\begin{table*}[ht] 
\centering
\renewcommand{\arraystretch}{1.2}
\scalebox{0.9}{
\begin{tabular}{@{}lccc@{}} 
\toprule
LLMs         & Method           & Average Tokens Processed (Tacred/Semeval)	&GPU Type      \\
\midrule
\multirow{4}{*}{{\parbox{2.28cm}{\textbf{Qwen2.5-7B-Instruct-1M}}}} &
Multi-Class & 1920/860&	NVIDIA RTX 3090*1	\\
& M-BRe 4 & 3560/2400&	NVIDIA RTX 3090*1	\\
& M-BRe $\lfloor N/6 \rfloor$ & 4655/2100	&NVIDIA RTX 3090*1\\
& Binary-Class & 14070/5700&	NVIDIA RTX 3090*1	\\
\midrule 
\multirow{4}{*}{{\parbox{2.28cm}{\textbf{Qwen2.5-14B-Instruct-1M}}}} &
Multi-Class & 1920/860&	NVIDIA RTX 3090*2	\\
& M-BRe 4 & 3560/2400	&NVIDIA RTX 3090*2	\\
& M-BRe $\lfloor N/6 \rfloor$ & 4655/2100&	NVIDIA RTX 3090*2\\
& Binary-Class & 14070/5700&	NVIDIA RTX 3090*2	\\  
\midrule 
\multirow{4}{*}{{\parbox{2.28cm}{\textbf{Qwen3-14B}}}} &
Multi-Class & 1920/860&	NVIDIA RTX 3090*2	\\
& M-BRe 4 & 3560/2400&	NVIDIA RTX 3090*2	\\
& M-BRe $\lfloor N/6 \rfloor$ & 4655/2100&	NVIDIA RTX 3090*2\\
& Binary-Class & 14070/5700&	NVIDIA RTX 3090*2	\\
\bottomrule
\end{tabular}}
\caption{\label{tab:resource consumption}Resource consumption of various LLMs under different frameworks.}
\end{table*}

\begin{table*}[t]

\centering
\renewcommand{\arraystretch}{1}
\scalebox{0.86}{
\begin{tabular}{ccccccc|cccccc}
  \toprule
  \multirow{2}*{Group} & \multicolumn{6}{c}{Qwen2.5-7B-Instruct-1M for Tacred} & \multicolumn{6}{c}{Qwen2.5-7B-Instruct-1M for SemEval} \\
  \cmidrule(r){2-7}\cmidrule(r){8-13}
  & Ma-P & Ma-R & Ma-F1 & Mi-F1 & S\_A\_F1 & \multicolumn{1}{c|}{Time(h)}
  & Ma-P & Ma-R & Ma-F1 & Mi-F1 & S\_A\_F1 & Time(h)\\
  \cmidrule(r){1-13}
  \multicolumn{1}{c|}{Mix2} & 44.44 & 54.49 & 44.52 & 45.51 & 38.81 & 0.22 & 34.07 & 37.59 & 32.63 & 32.08 & 28.94 & 0.09\\
  \multicolumn{1}{c|}{Mix3} & 50.24 & 68.79 & 50.24 & 52.20 & 42.11 & 0.33  & \textbf{40.66} & \textbf{47.06} & \textbf{38.95} & \textbf{39.52} & \textbf{32.05} & \textbf{0.13}\\
  \multicolumn{1}{c|}{Mix4} & \underline{54.11} & \underline{70.70} & \underline{54.05} & \underline{57.10} & \underline{42.19} & \underline{0.46}  & \underline{36.26} & \underline{35.25} & \underline{34.74} & \underline{30.98} & \underline{27.29} & \underline{0.16}\\
  \multicolumn{1}{c|}{Mix5} & 55.55 & 72.11 & 55.48 & 58.48 & 39.61 & 0.53  & 27.47 & 26.13 & 26.32 & 23.47 & 20.37 & 0.21\\
  \multicolumn{1}{c|}{Mix6} & 56.04 & 70.33 & 55.95 & 57.86 & 40.53 & 0.69  & 27.47 & 24.04 & 26.32 & 21.52 & 19.93 & 0.24\\
  \multicolumn{1}{c|}{Mix7} & \textbf{57.97} & \textbf{73.97} & \textbf{57.86} & \textbf{60.50} & \textbf{38.91} & \textbf{0.72} & 28.57 & 31.43 & 27.37 & 24.85 & 19.34 & 0.27\\
  \multicolumn{1}{c|}{Mix8} & 56.04 & 71.68 & 55.95 & 58.41 & 37.92 & 0.84 & 31.87 & 23.72 & 30.53 & 25.28 & 19.93 & 0.28\\
  \multicolumn{1}{c|}{Mix9} & 56.04 & 74.35 & 55.95 & 59.37 & 37.04 & 0.93 & 30.77 & 35.98 & 29.47 & 26.35 & 18.92 & 0.32\\
  \multicolumn{1}{c|}{Mix10} & 55.07 & 67.38 & 55.00 & 56.15 & 35.33 & 1.17 & - & - & - & - & - & -\\
  \multicolumn{1}{c|}{Mix11} & 57.00 & 74.66 & 56.90 & 59.74 & 35.47 & 1.18 & - & - & - & - & - & -\\
  \multicolumn{1}{c|}{Mix12} & 57.97 & 73.85 & 57.86 & 61.27 & 36.06 & 1.23 & - & - & - & - & - & -\\
  \multicolumn{1}{c|}{Mix13} & 57.97 & 73.60 & 57.86 & 61.51 & 33.89 & 1.49 & - & - & - & - & - & -\\
  \cmidrule(r){1-13}
  \multirow{2}*{Group} & \multicolumn{6}{c}{Qwen2.5-14B-Instruct-1M for Tacred} & \multicolumn{6}{c}{Qwen2.5-14B-Instruct-1M for SemEval} \\
  \cmidrule(r){2-7}\cmidrule(r){8-13}
  & Ma-P & Ma-R & Ma-F1 & Mi-F1 & S\_A\_F1 & \multicolumn{1}{c|}{Time(h)}
  & Ma-P & Ma-R & Ma-F1 & Mi-F1 & S\_A\_F1 & Time(h)\\
  \cmidrule(r){1-13}
  \multicolumn{1}{c|}{Mix2} & 57.00 & 68.32 & 56.90 & 57.75 & 51.37 & 0.87 & 49.45 & 53.44 & 47.37 & 46.19 & 40.29 & 0.27\\
  \multicolumn{1}{c|}{Mix3} & 56.52 & 68.03 & 56.43 & 58.19 & 45.49 & 1.27 & \textbf{54.95} & \textbf{55.75} & \textbf{56.84} & \textbf{52.75} & \textbf{39.56} & \textbf{0.42}\\
  \multicolumn{1}{c|}{Mix4} & \underline{59.90} & \underline{73.86} & \underline{59.76} & \underline{61.35} & \underline{45.22} & \underline{1.66} & \underline{57.14} & \underline{58.48} & \underline{54.74} & \underline{51.12} & \underline{37.07} & \underline{0.63}\\
  \multicolumn{1}{c|}{Mix5} & 59.42 & 69.12 & 59.29 & 59.76 & 41.47 & 2.26 & 59.34 & 65.61 & 61.05 & 59.25 & 38.72 & 0.71\\
  \multicolumn{1}{c|}{Mix6} & 62.32 & 81.09 & 62.14 & 65.37 & 41.76 & 3.02 & 61.54 & 54.50 & 58.95 & 54.93 & 35.90 & 0.73\\
  \multicolumn{1}{c|}{Mix7} & \textbf{63.29} & \textbf{77.79} & \textbf{63.10} & \textbf{65.55} & \textbf{38.27} & \textbf{3.08} & 63.74 & 61.05 & 61.05 & 58.41 & 39.35 & 0.87\\
  \multicolumn{1}{c|}{Mix8} & 63.29 & 80.42 & 63.10 & 65.09 & 37.15 & 3.77 & 63.74 & 72.58 & 61.05 & 61.22 & 30.97 & 1.03\\
  \multicolumn{1}{c|}{Mix9} & 63.77 & 84.45 & 63.57 & 67.65 & 34.96 & 3.87 & 64.84 & 77.98 & 66.32 & 65.91 & 27.35 & 1.18\\
  \multicolumn{1}{c|}{Mix10} & 62.80 & 80.64 & 62.62 & 65.88 & 31.26 & 4.27 & - & - & - & - & - & -\\
  \multicolumn{1}{c|}{Mix11} & 65.22 & 85.64 & 65.00 & 68.87 & 31.65 & 4.50 & - & - & - & - & - & -\\
  \multicolumn{1}{c|}{Mix12} & 64.25 & 84.97 & 64.05 & 67.80 & 30.23 & 5.57 & - & - & - & - & - & -\\
  \multicolumn{1}{c|}{Mix13} & 66.67 & 85.19 & 66.43 & 70.14 & 30.15 & 6.33 & - & - & - & - & - & -\\
  \cmidrule(r){1-13}
  \multirow{2}*{Group} & \multicolumn{6}{c}{Qwen3-14B for Tacred} & \multicolumn{6}{c}{Qwen3-14B for SemEval} \\
  \cmidrule(r){2-7}\cmidrule(r){8-13}
  & Ma-P & Ma-R & Ma-F1 & Mi-F1 & S\_A\_F1 & \multicolumn{1}{c|}{Time(h)}
  & Ma-P & Ma-R & Ma-F1 & Mi-F1 & S\_A\_F1 & Time(h)\\
  \cmidrule(r){1-13}
  \multicolumn{1}{c|}{Mix2} & 68.12 & 77.20 & 68.57 & 69.43 & 56.84 & 0.34 & 15.38 & 24.85 & 14.74 & 14.23 & 14.29 & 0.11\\
  \multicolumn{1}{c|}{Mix3} & 69.57 & 77.15 & 69.29 & 70.03 & 51.21 & 0.56 & \textbf{21.98} & \textbf{30.50} & \textbf{25.26} & \textbf{21.16} & \textbf{17.22} & \textbf{0.17}\\
  \multicolumn{1}{c|}{Mix4} & \underline{70.53} & \underline{76.91} & \underline{70.95} & \underline{70.68} & \underline{46.07} & \underline{0.61} & \underline{20.88} & \underline{28.06} & \underline{24.21} & \underline{19.76} & \underline{15.38} & \underline{0.25}\\
  \multicolumn{1}{c|}{Mix5} & 71.98 & 81.21 & 72.38 & 72.83 & 43.41 & 0.80 & 13.19 & 11.22 & 12.63 & 7.84 & 9.30 & 0.31\\
  \multicolumn{1}{c|}{Mix6} & 72.46 & 87.53 & 72.86 & 74.73 & 42.87 & 0.85 & 15.38 & 21.12 & 14.74 & 11.69 & 11.02 & 0.41\\
  \multicolumn{1}{c|}{Mix7} & \textbf{75.85} & \textbf{84.19} & \textbf{76.19} & \textbf{76.89} & \textbf{41.15}& \textbf{0.87} & 12.08 & 24.01 & 11.58 & 10.38 & 6.69 & 0.45\\
  \multicolumn{1}{c|}{Mix8} & 77.29 & 89.87 & 77.62 & 79.32 & 37.96& 0.99 & 10.99 & 17.00 & 10.53 & 8.27 & 5.07 & 0.51\\
  \multicolumn{1}{c|}{Mix9} & 75.85 & 86.52 & 76.19 & 77.74 & 35.46& 1.08 & 15.38 & 15.65 & 18.95 & 12.87 & 6.79 & 0.44\\
  \multicolumn{1}{c|}{Mix10} & 76.33 & 89.20 & 76.67 & 78.45 & 35.82& 1.23 & - & - & - & - & - & -\\
  \multicolumn{1}{c|}{Mix11} & 76.33 & 85.86 & 75.95 & 77.47 & 31.88& 1.28 & - & - & - & - & - & -\\
  \multicolumn{1}{c|}{Mix12} & 76.33 & 86.08 & 76.67 & 78.49 & 31.27& 1.73 & - & - & - & - & - & -\\
  \multicolumn{1}{c|}{Mix13} & 77.29 & 88.28 & 77.62 & 79.25 & 30.42& 1.75& - & - & - & - & - & -\\
  \bottomrule
\end{tabular}}
\caption{\label{tab:multi-binary}
Comprehensive assessment of different Number of Groupings. \textbf{Ma-P}, \textbf{Ma-R}, \textbf{Ma-F1}, \textbf{Mi-F1} and \textbf{S\_A\_F1} respectively represent Macro Precision, Macro Recall, Macro F1, Micro F1 and Special\_Avg\_F1. \textbf{Bold} denotes the optimal trade-off point.
}
\end{table*}

\begin{table*}

\centering
\renewcommand{\arraystretch}{1.2}
\begin{tabularx}{\textwidth}{l|>{\arraybackslash}X}
    \hline
    \textbf{Relation} & \textbf{Explanation}\\
    \hline
    Component-Whole (e2,e1)& Tail entity e2 is the component of head entity e1, and head entity e1 is the whole of tail entity e2. \\   
    \hline
    Instrument-Agency (e2,e1)	&Tail entity e2 is the instrument of head entity e1, and head entity e1 is the agency of tail entity e2.\\
    \hline
    Member-Collection (e1,e2)	&Head entity e1 is the member of tail entity e2, and tail entity e2 is the collection of head entity e1.\\
    \hline
    Cause-Effect (e2,e1)&	Tail entity e2 is the cause of head entity e1, and head entity e1 is the effect of tail entity e2.\\
    \hline
    Entity-Destination (e1,e2)&	Head entity e1 is the entity of tail entity e2, and tail entity e2 is the destination of head entity e1.\\
    \hline
    Content-Container (e1,e2)	&Head entity e1 is the content of tail entity e2, and tail entity e2 is the container of head entity e1.\\
    \hline
    Message-Topic (e1,e2)&	Head entity e1 is the message of tail entity e2, and tail entity e2 is the topic of head entity e1.\\
    \hline
    Product-Producer (e2,e1)&	Tail entity e2 is the product of head entity e1, and head entity e1 is the producer of tail entity e2.\\
    \hline
    Member-Collection (e2,e1)&	Tail entity e2 is the member of head entity e1, and head entity e1 is the collection of tail entity e2.\\
    \hline
    Entity-Origin (e1,e2)	&Head entity e1 is the entity of tail entity e2, and tail entity e2 is the origin of head entity e1.\\
    \hline
    Cause-Effect (e1,e2)	&Head entity e1 is the cause of tail entity e2, and tail entity e2 is the effect of head entity e1.\\
    \hline
    Component-Whole (e1,e2)&	Head entity e1 is the component of tail entity e2, and tail entity e2 is the whole of head entity e1.\\
    \hline
    Message-Topic (e2,e1)&	Tail entity e2 is the message of head entity e1, and head entity e1 is the topic of tail entity e2.\\
    \hline
    Product-Producer (e1,e2)&	Head entity e1 is the product of tail entity e2, and tail entity e2 is the producer of head entity e1.\\
    \hline
    Entity-Origin (e2,e1)	&Tail entity e2 is the entity of head entity e1, and head entity e1 is the origin of tail entity e2.\\
    \hline
    Content-Container (e2,e1)&	Tail entity e2 is the content of head entity e1, and head entity e1 is the container of tail entity e2.\\
    \hline
    Instrument-Agency (e1,e2)&	Head entity e1 is the instrument of tail entity e2, and tail entity e2 is the agency of head entity e1.\\
    \hline
    Entity-Destination (e2,e1)	&Tail entity e2 is the entity of head entity e1, and head entity e1 is the destination of tail entity e2.\\
    \hline
    Other	&There is no relationship or unrecognized relationship between the head and tail entities.\\
    \hline
    org:founded&	The founding time of an organization.\\
    \hline
    org:subsidiaries	&The subsidiaries of an organization.\\
    \hline
    per:date\_of\_birth&	The date of birth of a person.\\
    \hline
    per:cause\_of\_death	&The cause of death of a person.\\
    \hline
    per:age	&The age of a person.\\
    \hline
    per:stateorprovince\_of\_birth&	The state or province of birth of a person.\\
    \hline
\end{tabularx}
\end{table*}

\begin{table*}
\centering
\renewcommand{\arraystretch}{1.2}
\begin{tabularx}{1.08\textwidth}{l|>{\arraybackslash}X}
    \hline
    \textbf{Relation} & \textbf{Explanation}\\
    \hline
    per:countries\_of\_residence	&The countries where a person resides.\\
    \hline
    per:country\_of\_birth&	The country of birth of a person.\\
    \hline
    per:stateorprovinces\_of\_residence	&The states or provinces where a person resides.\\
    \hline
    org:website&	The website of an organization.\\
    \hline
    per:cities\_of\_residence&	The cities where a person resides.\\
    \hline
    per:parents	&The parents of a person.\\
    \hline
    per:employee\_of	&The organization where a person is employed.\\
    \hline
    NA/no\_relation	&Unknown or non-existent relation.\\
    \hline
    per:city\_of\_birth	&The city of birth of a person.\\
    \hline
    org:parents	&The parent company of an organization.\\
    \hline
    org:political/religious\_affiliation&	The political or religious affiliation of an organization.\\
    \hline
    per:schools\_attended&	The schools attended by a person.\\
    \hline
    per:country\_of\_death	&The country where a person died.\\
    \hline
    per:children&	The children of a person.\\
    \hline
    org:top\_members/employees&	The top members/employees of an organization.\\
    \hline
    per:date\_of\_death	&The date of death of a person.\\
    \hline
    org:members&	The members of an organization.\\
    \hline
    org:alternate\_names&	The alternate names of an organization.\\
    \hline
    per:religion	&The religion of a person.\\
    \hline
    org:member\_of&	The organization to which a member belongs.\\
    \hline
    org:city\_of\_headquarters	&The city where the headquarters of an organization is located.\\
    \hline
    per:origin	&The origin of a person.\\
    \hline
    org:shareholders	&The shareholders of an organization.\\
    \hline
    per:charges&	The charges against a person.\\
    \hline
    per:title&	The occupation of a person.\\
    \hline  
    org:number\_of\_employees/members&	The number of employees/members in an organization.\\
    \hline
    org:dissolved	&The date of dissolution of the organization.\\
    \hline
    org:country\_of\_headquarters	&The country where headquarters of an organization is located.\\
    \hline
    per:alternate\_names	&The alternate names of a person.\\
    \hline
    per:siblings	&The siblings of a person.\\
    \hline
    org:stateorprovince\_of\_headquarters&	The state or province where headquarters of an organization is located.\\
    \hline
    per:spouse	&The spouse of a person.\\
    \hline
    per:other\_family	&Other family members of a person.\\
    \hline
    per:city\_of\_death	&The city where a person died.\\
    \hline
    per:stateorprovince\_of\_death&	The state or province where a person died.\\
    \hline
    org:founded\_by&	The founder of an organization.\\
    \hline
    org:country\_of\_branch&	The country where a branch of an organization is located.\\
    \hline
    org:city\_of\_branch	&The city where a branch of an organization is located.\\
    \hline
    org:stateorprovince\_of\_branch&	The state or province where branch of an organization is located.\\
    \hline
    per:identity	&The identity information or characteristics of a person.\\
    \hline
\end{tabularx}
\caption{\label{tab:Relation Explanation}
Explanation of each relation in the four datasets.  
}
\end{table*}
\end{document}